\documentclass[lettersize,journal]{IEEEtran}
\usepackage{amsmath,amsfonts}
\usepackage{algorithmic}
\usepackage{algorithm}
\usepackage{array}
\usepackage{textcomp}
\usepackage{stfloats}
\usepackage{url}
\usepackage{verbatim}
\usepackage{graphicx}
\usepackage{cite}
\usepackage{multirow}
\usepackage[table,xcdraw]{xcolor}
\usepackage[normalem]{ulem}
\useunder{\uline}{\ul}{}
\usepackage{subcaption}  % 添加 subcaption 宏包
\usepackage{bm}

\newtheorem{definition}{Definition}
\newtheorem{problem}{Problem}
\newcommand{\tabincell}[2]{\begin{tabular}{@{}#1@{}}#2\end{tabular}}
\begin{document}
\begin{sloppypar}

\title{FairSTG: Countering performance heterogeneity via collaborative sample-level optimization}

\author{
Gengyu Lin, Zhengyang Zhou{*}~\IEEEmembership{Member,~IEEE}, Qihe Huang, Kuo Yang,\\ Shifen Cheng, Yang Wang{*}~\IEEEmembership{Senior Member,~IEEE}
        % <-this % stops a space
\thanks{G Lin,  Q Huang, K Yang are  with University of Science and Technology of China (USTC). 
 Y Wang and Z Zhou are joint corresponding authors, and they are with both Suzhou Institute for Advanced Research, USTC, and USTC. Z Zhou  is also with State Key Laboratory of Resources and EnvironmentalInformation System. S Cheng is with Institute of Geographic Sciences and Natural Resources Research, Chinese Academy of Sciences, and University of Chinese Academy of Sciences.
 }
\thanks{Email: \{lingengyu, hqh, yangkuo\}@mail.ustc.edu.cn,
\{angyan, zzy0929\}@ustc.edu.cn, chengsf@lreis.ac.cn}% <-this % stops a space
\thanks{Manuscript received Feb 1, 2024; revised xxx xx, 2024.}}

\markboth{Journal of \LaTeX\ Class Files,~Vol.~14, No.~8, August~2021}%
{Shell \MakeLowercase{\textit{et al.}}: A Sample Article Using IEEEtran.cls for IEEE Journals}



\maketitle

\begin{abstract}
Spatiotemporal learning plays a crucial role in mobile computing techniques to empower smart cites. While existing research has made great efforts to achieve accurate predictions on the overall dataset, they still neglect the significant performance heterogeneity across samples. In this work, we designate the performance heterogeneity as the reason for unfair spatiotemporal learning, which not only degrades the practical functions of models, but also brings serious potential risks to real-world urban applications. To fix this gap, we propose a model-independent Fairness-aware framework for SpatioTemporal Graph learning (FairSTG), which inherits the  idea of  exploiting advantages of  well-learned  samples to challenging ones with collaborative mix-up. Specifically, FairSTG consists of a spatiotemporal feature extractor for model initialization, a collaborative representation enhancement for knowledge transfer  between well-learned samples and challenging ones, and fairness objectives for immediately suppressing sample-level performance heterogeneity. Experiments on four spatiotemporal datasets demonstrate that our FairSTG significantly improves the fairness quality while maintaining comparable forecasting accuracy. Case studies show FairSTG can counter both spatial and temporal performance heterogeneity by our sample-level retrieval and compensation, and our work can potentially alleviate the risks on spatiotemporal resource allocation for underrepresented urban regions.

\end{abstract}


\section{Introduction}

\IEEEPARstart{W}{ith} the rapid urbanization and increasing number of urban devices, we are now embracing a new era with a vast amount of valuable spatiotemporal data. Actually, spatiotemporal data plays a crucial role in  mobile computing services in smart cities, including traffic police assignment~\cite{ji2023spatio, bai2020adaptive, li2017diffusion}, urban safety management~\cite{zhou2020foresee, an2022hintnet, zheng2023deep}, and numerical weather forecasting ~\cite{shang2021discrete}. However, data collected from real-world is inevitably trapped into bias due to imbalance sampling, inherent low quality or under-representation in gender, race or other sensitive attributes.

Recently,  fairness issue, which calls for the equal opportunity on allocation and assignment, has received increasing attention in machine learning, from recommendation system~\cite{morik2020controlling, wu2021learning, wang2022improving} to resource allocation~\cite{guo2023fairness}. Without explicitly considering the fairness issue,  machine learning models will erroneously learn such bias and even exacerbate the unfairness, leading to misleading decisions on downstream tasks~\cite{du2020fairness, mehrabi2021survey}.

\begin{table}[htbp]
\begin{center}
\caption{The unfairness issue in spatiotemporal learning. }
%Although state-of-the-art methods achieve superior overall performance, the performance heterogeneity is non-negligible.
\label{table:unfairness}
\begin{tabular}{lllll}
\hline
                          &               & DCRNN   & MTGNN  & D$^2$STGNN \\ \hline
\multirow{4}{*}{METR-LA}  & overall MAE   & 3.57    & 3.49   & 3.35    \\
                          & MAE variance  & 51.61  & 47.68  & 49.47   \\
                          & overall MAPE  & 10.40\% & 9.87\% & 9.43\%  \\
                          & MAPE variance & 0.18    & 0.15   & 0.14    \\ \hline
\multirow{4}{*}{PEMS-BAY} & overall MAE   & 1.96    & 1.96   & 1.98    \\
                          & MAE variance  & 17.23   & 16.50  & 17.02   \\
                          & overall MAPE  & 4.64\%  & 4.67\% & 4.84\%  \\
                          & MAPE variance & 0.05    & 0.04   & 0.06    \\ \hline
\end{tabular}
\end{center}
\end{table}

\begin{figure}[htbp]
    \centering
    \includegraphics[width=\linewidth]{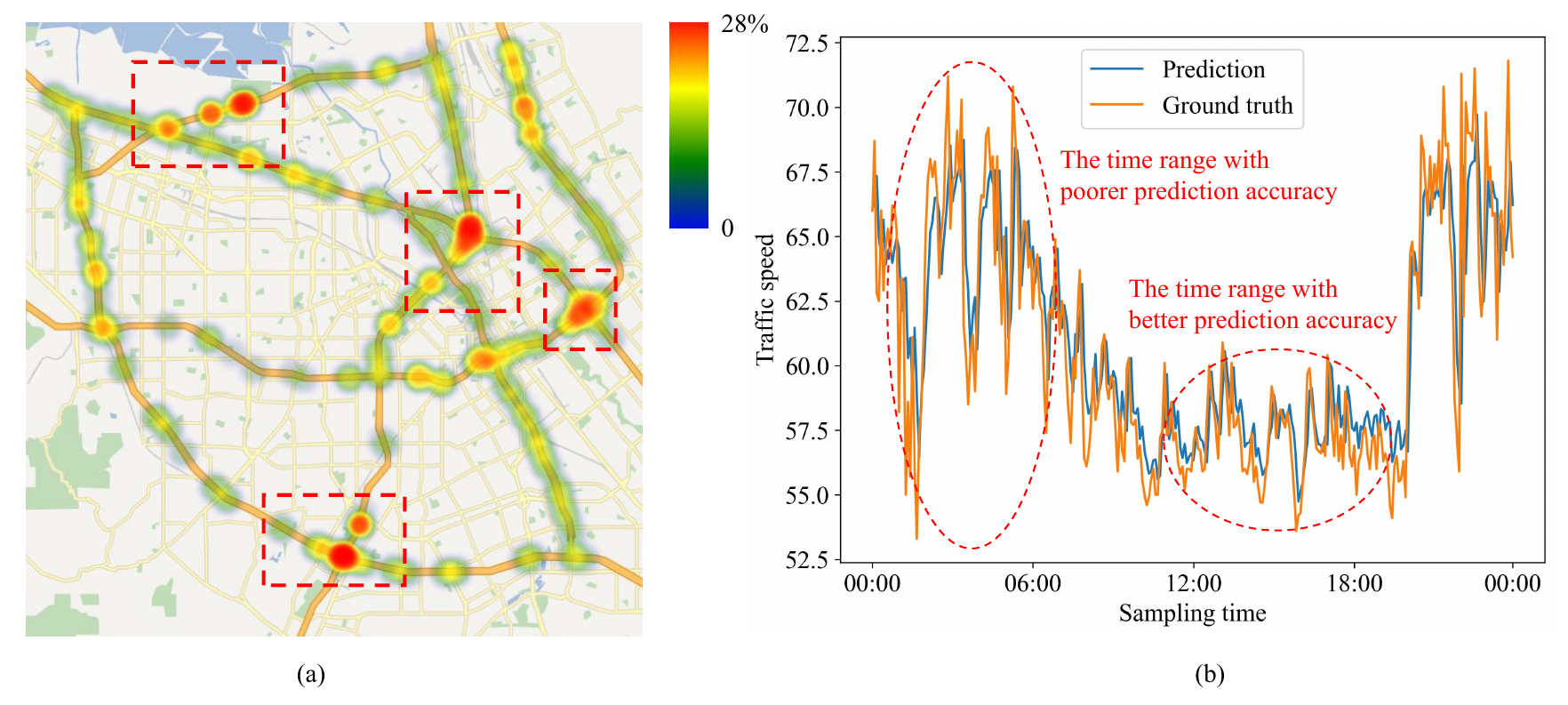}
    \caption{The unfairness across spatial and temporal ranges. (a) The mean MAPE of each sensor in PEMS-BAY dataset generated by MTGNN, with red boxes indicating regions with significant errors. (b) The curve shows the traffic flow data and the corresponding predictions from MTGNN on sensor \#64 in PEMS-BAY dataset. It can be observed that the prediction performance varies significantly at different time stamps even for the same sensor.}
    \label{fig:unfair_st}
\end{figure}

In fact, in spatiotemporal learning for mobile computing services, the unfairness phenomenon is even non-negligible. Current literature on spatiotemporal learning mostly concentrates on the overall performance, overlooking the performance heterogeneity across different samples and regions. 
A preliminary experiment conducted  on two well-known datasets verifies such serious performance heterogeneity. Consider the expectation of Mean Absolute Error (MAE)  and variance of MAE as the indicator for overall performance and sample-level  performance heterogeneity, we find that advanced spatiotemporal learning methods achieve satisfactory overall  performance, but the MAE variance is approximately 14 times of MAE on METR-LA and 9 times of MAE on PEMS-BAY (refer to Table~~\ref{table:unfairness}). Additionally, it is observed that this unfairness simultaneously exists in both the temporal and spatial domains. As shown in Fig. \ref{fig:unfair_st}(a), the sensors near transportation hubs exhibit poorer forecasting performance, possibly due to the more complex traffic conditions, while in Fig. \ref{fig:unfair_st}(b), the prediction performance varies significantly at different time stamps for the same sensor. We designate such prediction disparities as performance heterogeneity, which can be the intrinsic reason for unfairness in spatiotemporal forecasting.

%An experiment on METR-LA and PEMS-BAY further demonstrates the sample-level unfairness phenomenon in spatiotemporal learning. As illustrated in Table ~\ref{table:unfairness}, we let variance of errors be the sample-level disparity indicator on performances. It is observed that the advanced spatiotemporal learning method achieves overall satisfactory performance, but exhibit significant performance disparity among different samples. We define the performance heterogeneity among different samples, which manifests performance disparity, as the unfairness metric in spatiotemporal learning. 

Unfortunately, an unfair spatiotemporal model can induce risks on two aspects. First, overlooking critical information on underrepresented parts can lead to the failure of key event prediction such as accidents, and environmental crisis, thus bringing in safety risks. Second, in a technical perspective, those regions with high regularity are easy to learn and dominate the training process, which results in the functional degradation of overall evaluation metrics. To this end, a fair spatiotemporal learning framework is essential to empower non-biased urban applications. In this work, we dissect the inherent factors behind unfairness and counter the heterogeneity of prediction results across spatiotemporal domain to increase the quality of urban decisions, thus deducing the system risks, especially on underrepresented groups and individuals. 

Concretely, the performance heterogeneity in spatiotemporal learning can be attributed to two aspects, i.e., the adequacy of data representation and the inherent regularity within datasets. First, urban sensors are more concentrated on city centers than suburban areas, leading to inadequate representation of marginal areas in the overall dataset and increasing their learning difficulty. Second, different samples and regions exhibit diverse patterns, inducing different degrees of learning difficulty. Meanwhile, abundant local contexts in spatiotemporal data, such as geographical locations, functional regions, and temporal contexts, influence the data regularity in a complex and intertwined manner. To this end, we posit that samples with lower regularity and high learning difficulty pose a greater challenge for the model, and exhibit poorer predictive performance. Thus, improving the forecasting performance on these samples is crucial for countering the unfairness issue in spatiotemporal data.

Existing literature concerning this work can be summarized as two lines. The line of spatiotemporal learning takes efforts to model spatial and temporal heterogeneity on observations, achieving personalized node-level directional aggregation~\cite{zhou2023predicting, zhou2022greto}, but has never modeled the heterogeneity of prediction results, i.e., the prediction disparities  on regions and temporal steps, directly resulting in prediction unfairness. While the line of fair machine learning investigates how to separate the influence of sensitive factors during training, and design various fairness  objectives including inter-group and intra-group equality~\cite{he2023learning}. But all these techniques still have not been advanced to urban spatiotemporal learning. To this end, we argue that there are two specific challenges in constructing a fairness-aware spatiotemporal learning framework.

\begin{itemize} %
    \item Given that spatiotemporal data is equipped with complex and heterogeneous dependencies but lacking explicit sensitive factors, the first challenge is how to exploit the implicit factors to accurately identify the specific challenging samples suffering unfair performances. 
    \item On model design aspect, how to devise fairness-aware learning strategies and maximally exploit the available contexts and high-quality spatiotemporal representations to collaboratively enhance learning of challenging samples, becomes the second challenge. 
\end{itemize}

To address above challenges, in this work, we propose a model-independent \underline{Fair}ness-aware \underline{S}patio\underline{T}emporal \underline{G}raph learning (FairSTG) to counter the spatiotemporal performance heterogeneity for fair learning. Our FairSTG takes series of each region as the minimal sample unit but collaboratively optimizes the spatiotemporal graph representation with an integrated framework in a holistic manner, which allows location-based fairness and joint enhancement. Our collaboration can be interpreted  hierarchically as two aspects. For the whole framework, we optimize the fair-aware representations with intrinsic factor of representation collaboration and immediate factor of fairness-aware learning objectives. In the representation enhancement, we argue that samples those are hard to learn primarily hinder the fair prediction results. To this end, we design an auxiliary self-supervised task to actively identify challenging samples, and construct the compensatory sample sets for adaptive representation mix-up, exploiting advantages of well-learned superior representations to improve challenging samples. Regarding learning objectives, we construct a variance-based objective to directly force the model to optimize towards consistent performance, and take the variance-based performance evaluation to further remedy the function degeneration of pure error metrics. The contributions can be summarized as below. 
\begin{itemize} %有序列表
    \item This is the first effort that summarizes the serious outcomes of  spatiotemporal learning without fairness, and subsequently attributes such performance heterogeneity into defective objectives and inherent heterogeneity of data learning difficulty.
    \item We propose a novel fair mobile computing technique FairSTG, from both immediate and intrinsic perspectives, advancing modeling observation heterogeneity to performance heterogeneity. We minimize the variances across samples and improve prediction performance of the challenging samples by drawing common patterns from the well-matched and high-quality representations.
    \item We design a fairness metric adapting to spatiotemporal forecasting, which jointly evaluate learning frameworks with error-based metrics. Experiments show that our solution significantly improves the equality across performances, and achieves comparable or better accuracy against baselines. Case studies demonstrate that FairSTG can alleviate the risks on urban resource allocation for underrepresented urban regions, and FairSTG can potentially become a paradigm of fair urban computing for sustainable  urban development.
\end{itemize}

\section{Related Work}

\subsection{Spatiotemporal learning}

Spatiotemporal learning is a crucial technique to empower urban applications. In an early stage, researchers take spatiotemporal forecasting tasks as time-series predictions, and introduce statistical solutions such as ARIMA~\cite{box2015time}, VAR~\cite{geweke1977dynamic} to achieve forecasting. These statistical approaches are easy to implement with reasonable interpretability, but fail to simultaneously capture both spatial and temporal dependencies. With the prosperity of deep learning and Graph Neural Networks (GNNs), deep GNNs naturally have the edge on learning non-Euclidean spatial data and can potentially accommodate to various spatiotemporal learning tasks~\cite{jin2023spatio, yu2018spatio, wu2019graph}. In fact, spatiotemporal heterogeneity, which refers to the varying patterns across different temporal or spatial ranges, has been widely recognized in such data and raised more attention in academia~\cite{jin2023spatio}. Specifically, ST-SSL~\cite{ji2023spatio} emphasizes the heterogeneity across temporal steps and spatial regions with a pair-wise learning in a self-supervised manner, while HA-STGN~\cite{xu2023learning} is proposed by designing a direction-aware road network and imposing a time-aware graph attention mechanism to capture such heterogeneity. Even though existing works have explicitly taken the heterogeneous observations into account with different aggregation and representation strategies, the performance heterogeneity along both spatial and temporal dimensions has never been considered. We designate such imbalanced prediction performance as the prediction unfairness, and counter such issue in this paper.

\subsection{Fairness-aware machine learning}

The unfairness in machine learning, which may do harm to interests of a specific group or individual, has received extensive attention. The unfairness can primarily stem from the inherent data imbalance, and the unawareness of fairness in machine learning algorithms can further aggravate such bias, leading to severe unfair resource allocation and discrimination.

Plenty of literature has attempted to counteract the bias in both data and algorithms to achieve fair machine learning. Based three stages in machine learning systems, fairness-aware learning can be generally classified into three categories, methods during pre-processing, in-processing, and post-processing. First, the pre-processing solutions remove the underlying bias by augmenting and adjusting the training data, and then train a model on debiased data, where it can be considered as the data-aspect solution. Second, in-processing methods try to incorporate fairness metrics or propose adversarial learning objectives to obtain fair representations, which can be viewed as the model aspect paradigm to balance the accuracy and fairness during the learning process. For example, to protect the interest of minority news providers, ProFairRec~\cite{qi2022profairrec} exploits a sensitive attribute discriminator to identify the provider-bias information while generators make indistinguishable provider-fair representations against the discriminator. And VFAE~\cite{louizos2015variational} investigates a variational autoencoder with Maximum Mean Discrepancy to generate regularized fair representations. The post-processing methods allow transformations on model outputs, such as label re-assignment~\cite{hardt2016equality, pleiss2017fairness}, re-ranking of output lists~\cite{morik2020controlling, liu2019personalized, sonboli2020opportunistic} and projection of representations onto debiased subspace~\cite{ravfogel2020null, bolukbasi2016man}, to mitigate unfairness.

Even fair learning systems have been widely investigated, the unfairness issue in spatiotemporal learning, which directly leads to unfair resource allocation and underestimated risks, has still been under-explored. Such under-exploration can be attributed to two challenges. First, spatiotemporal data lacks explicit sensitive information for fairness constraints. Second, this kind of data exhibits complex spatial and temporal heterogeneity and dynamic variations~\cite{ji2023spatio}, contributing to the difficulty in  capturing real statuses of inactive spatial regions and interactions between active and inactive regions.

\section{Problem Formulation}

We focus on countering the unfairness issue in spatiotemporal learning while maintaining the performance of the backbone model.

% \noindent \textbf{Definition 1: Spatiotemporal graph. }
\begin{definition}[Spatiotemporal graph] 
A spatiotemporal graph (ST Graph) is formulated as $\mathbb{G} = \{ \mathcal{G}_1, \mathcal{G}_2, \cdot \cdot \cdot, \mathcal{G}_T\}$, to describe spatiotemporal data, where $\mathcal{G}=(\mathcal{V}, \mathcal{E}, \mathbf{A}, \mathbf{X})$. The node set $\mathcal{V}=\{v_1, v_2, \cdot \cdot \cdot, v_N\}$ and edge set $\mathcal{E} = \{e_{ij} = (v_i, v_j)\}$ can be formulated respectively,  where $N = |\mathcal{V}|$ is the number of nodes. It is worth noting that we do not force a predefined adjacency matrix for spatiotemporal graphs in our tasks, but it can be learned from input features. We then define $\mathbf{A} \in \mathbb{R}^{N \times N}$ as the virtual (learnable) adjacency matrix of our spatiotemporal graph when mentioned.
\end{definition}

Besides, let $\mathbf{X}_{:,0:T-1}=\{\mathbf{X}_{:,0}, \mathbf{X}_{:,1}, \cdot \cdot \cdot, \mathbf{X}_{:,T-1}\} \in \mathbb{R}^{N \times T}$ denote a series of observed spatiotemporal graphs with $N$ nodes and $T$ time steps, where $\mathbf{X}_{:,t}=\{x_{0,t}, x_{1,t}, \cdot \cdot \cdot, x_{N-1,t}\} \in \mathbb{R}^N$ records the observations of these $N$ nodes in $\mathcal{G}_t$ at time step $t$ and $x_{i,t}$ represents the deterministic value of node $i$ at time step $t$. 

% \noindent \textbf{Problem 1: Spatiotemporal forecasting.}
\begin{problem}[Spatiotemporal forecasting] 
Given $\mathbf{X}_{:,t-w:t-1} \in \mathbb{R}^{N \times w}$, spatiotemporal forecasting aims to derive the following $h$ steps of observations. 
%$Y = X_{:,t:t+h-1}=\{X_{:,t}, X_{:,t+1}, ..., X_{:, t+h-1}\}  \in \mathbb{R}^{N \times h}$%
Then, the spatiotemporal prediction problem can be formalized as follows,
\begin{equation}
    \widehat{\mathbf{Y}}_{:,t:t+h-1} = f_{\theta}(\mathbf{X}_{:,t-w:t-1};\mathcal{G})
\end{equation}
where $f$ represents the model for spatiotemporal prediction, and $\theta \in \Theta$ denotes the learnable parameters.

We refer to the observations of the $i$-th node with a temporal window of length $w$ before time step $t$ as a \textbf{spatiotemporal sample}, formalized as $\mathbf{X}_{i,t-w:t-1}$. And the corresponding ground truth is defined as $\mathbf{X}_{i,t:t+h-1}$. For ease of description, we denote the set of input spatiotemporal samples as $\mathcal{X} = \{\bm{x}_1, \bm{x}_2, \cdot \cdot \cdot, \bm{x}_M\}$, where $\bm{x}_i \in \mathbb{R}^w$ represents an individual sample and $M=|\mathcal{X}|$ denotes the number of samples. And the set of ground truth is formulated as $\mathcal{Y}=\{\bm{y}_1, \bm{y}_2, \cdot \cdot \cdot, \bm{y}_M\}$, where $\bm{y}_i \in \mathbb{R}^h$. Then we can explicitly capture the spatial heterogeneity via our node-level defined spatiotemporal sample. Given above, the spatiotemporal forecasting problem can be reformulated as,
$$
\hat{\mathcal{Y}} = f_{\theta}(\mathcal{X})
$$

In this work, we take MAE as the error metric, and the objective of the forecasting task is to find an optimal set of parameters $\theta^* \in \Theta$ that minimizes the global MAE, i.e.,
\begin{equation}
\begin{aligned}
  \theta^* &= \arg_{\theta^* \in \Theta} \min \text{MAE}(\mathcal{Y}, \hat{\mathcal{Y}}) \\
  &= \arg_{\theta^* \in \Theta} \min \text{MAE}(\mathcal{Y}, f_{\theta^*}(\mathcal{X}))  
\end{aligned}
\end{equation}
\end{problem}

%Given the fixed input length $w \in \mathbb{N}^+$ and output length $h \in \mathbb{N}^+$, the input of the spatiotemporal forecasting is formulated as $X_{:,t-w:t-1}=\{X_{:,t-w}, X_{:,t-w+1}, ..., X_{:,t-1}\} \in \mathbb{R}^{N \times w}$, and $Y_{:,t-w:t-1} \in \mathbb{R}^{N \times h}$ denotes the prediction of the forecasting model. The corresponding ground truth is formulated as $X_{:,t:t+h-1}=\{X_{:,t}, X_{:,t+1}, ..., X_{:, t+h-1}\} \in \mathbb{R}^{N \times h}$. The spatiotemporal prediction problem can be formalized as follows:

% We refer to the observations of the $i$-th node with a temporal window of length $w$ before time step $t$ as a spatiotemporal sample, formalized as $X_{i;t-w:t-1}$. The corresponding prediction is defined as $Y_{i,t:t+h-1}=f(X_{i,t-w:t-1};\mathcal{G})$ and the ground truth is defined as $X_{i,t:t+h-1}$. For ease of description, we denote the set of input spatiotemporal samples as $X = \{x_1, x_2, ..., x_M\}$, where $x_i \in \mathbb{R}^w$ represents an individual sample and $M=|X|$ denotes the number of nodes. And the ground truth of output is formulated as $Y=\{y_1, y_2, ..., y_M\}$, where $y_i \in \mathbb{R}^h$. Then the spatiotemporal forecasting problem can be reformulated as:
% \begin{equation}
%     \hat{Y} = f_{\theta}(X)
% \end{equation}

% \noindent \textbf{Definition 2:  Fairness metrics in spatiotemporal forecasting.}
\begin{definition}[Fairness metrics in spatiotemporal forecasting]
% We utilize the relative error between predictions and ground truth to quantify the prediction error of an individual sample: 
% \begin{equation}
%     f_{err}(y_i, \hat{y_i}; \theta) = |\frac{\hat{y_i}-y_i}{y_i}|
% \end{equation}
We posit that a fair spatiotemporal prediction model should provide predictions with similar performance for different spatiotemporal samples, meaning that the prediction errors for various samples are close. We take \textbf{the disparity in errors among different spatiotemporal samples} to charaterize the degree of the performance unfairness. And we employ \textbf{the variance of errors among different spatiotemporal samples} to quantify the disparity, illustrated as $D(\mathcal{Y},\hat{\mathcal{Y}};\theta)$, where multiple metrics can be chosen to measure  prediction errors. The larger the variance is, the more unfair the forecasts are. 
\end{definition}
% The discrepancy can by formulated as follows:
% \begin{equation}
%     \text{Disparity}(Y,\hat{Y};\theta) = \frac{1}{M}\sum_{i=0}^{M-1}[f_{err}(y_i, \hat{y_i};\theta)-  
%     \frac{1}{M}\sum_{j=0}^{M-1}f_{err}(y_j, \hat{y_j};\theta)]^2
% \end{equation}

% \noindent \textbf{Problem 2: Fairness-aware spatiotemporal forecasting.}
\begin{problem}[Fairness-aware spatiotemporal forecasting]
In this work, we improve the spatiotemporal forecasting model $f_{\theta^*}$ to a fairness-aware version, which enforces the model to treat different spatiotemporal samples fairly and  simultaneously maintais the original prediction performance. It can be formally formulated as,
\begin{equation}
\begin{aligned}
\tilde\theta &= \arg_{\tilde\theta \in \Theta} \min D(\mathcal{Y},\hat{\mathcal{Y}};\tilde\theta) \\ 
\text{s.t.} &\ \text{MAE}(\mathcal{Y}, f_{\tilde\theta}(\mathcal{X})) \le \delta \cdot \text{MAE}(\mathcal{Y}, f_{\theta^*}(\mathcal{X}))
\end{aligned}    
\end{equation}
Here, $\delta \textgreater 0$ describes the trade-off between mitigating unfairness and the inevitable sacrifice in overall performance of the spatiotemporal prediction model.
It also serves as an evaluation metric, measuring the overall performance sacrifice incurred by the alleviating the unfairness issue.
If $\delta > 1$, it indicates that the model sacrifices some accuracy to ensure global fair predictions. If $0 < \delta \le 1$, it suggests that the model simultaneously improves both fairness and forecasting performance.
\end{problem}

\section{Methodology}
\begin{figure*}[htbp]
    \centering
    \includegraphics[width=\textwidth]{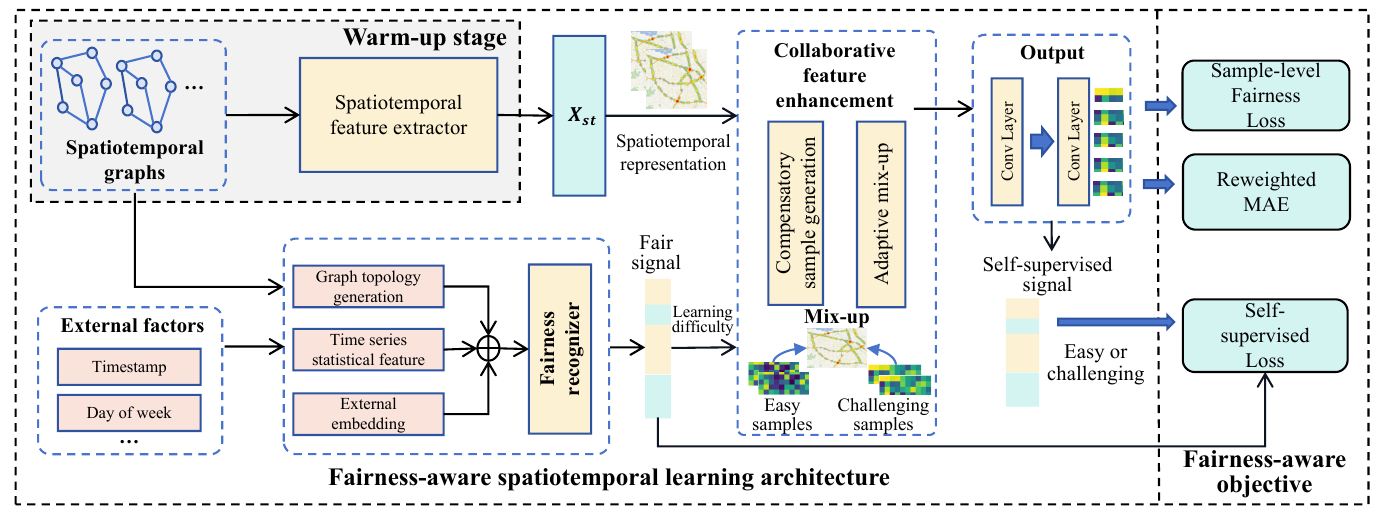}
    \caption{Framework  overview of FairSTG. The spatiotemporal feature extractor learns spatiotemporal representations from the original ST Graph. The fairness recognizer mines the learning difficulty and generates fairness signals in a self-supervised manner, and the collaborative feature  enhancement adaptively transfers advantageous features from easy set to challenging set. Finally, the output module transforms the fused representations and produces the predictions.}
    \label{fig:framework}
\end{figure*}

To counter the unfairness issue in spatiotemporal forecasting, we propose a novel \underline{Fair}ness-aware \underline{S}patio\underline{T}emporal \underline{G}raph learning (FairSTG), which ensures fair global predictions via collaborative feature transfer and fairness constraints. As illustrated in Figure ~\ref{fig:framework}, FairSTG consists of four well-designed components, spatiotemporal feature extractor, fairness recognizer, collaborative feature enhancement and output module.
% the warm-up stage with spatiotemporal feature extractor (ST extractor), and the fairness learning stage, with collaborative feature enhancement and spatiotemporal fairness-aware objective. 
To be specific, spatiotemporal feature extractor generates the spatiotemporal representations from the input ST Graph. Fairness recognizer identifies the learning difficulty of samples and generates fairness signals in a self-supervised manner. Then, based on the fairness signals, we propose a collaborative feature enhancement to adaptively improve the informativeness of representations via transferring the advantageous representations from well-learned samples to those difficult to learn. Finally, we output fair forecasts at one forward step. 

%Fairness-aware module and feature synergetic enhancement module work cooperatively, perceiving the learning difficulty of various samples and compensating the spatiotemporal samples that are prone to unfair treatment by the model, to proactively ensure fairness in global predictions. Simultaneously, we introduce fairness metrics into our objective, designing a dual mechanism at both representation-level and objective-level to insure fair spatiotemporal forecasting.

\subsection{Spatiotemporal feature extractor}

% The spatiotemporal feature extractor is also considered as the warm-up stage to initialize the learnable parameters. 
The spatiotemporal feature extractor captures sequential patterns and spatial correlations from the input ST Graph. Existing literature designs various models for feature extraction based on specific application scenarios, such as RNN-based, CNN-based and GNN-based models. Our FairSTG is a model-independent framework, allowing different models to work as the backbone for the extractor, according to different application scenarios and data characteristics. In our work, we select two superior spatiotemporal GNN-based models MTGNN~\cite{wu2020connecting} and D$^2$STGNN~\cite{shao2022decoupled} as the backbone model of the extractor. MTGNN designs a graph learning layer to extract uni-directional relations among nodes, and a novel mix-hop propagation layer and a dilated inception layer are further proposed to capture the spatial and temporal dependencies within the time series. D$^2$STGNN decouples and handles the diffusion and inherent traffic information separately. Note that in our work, the backbone can be easily replaced with any other prevalent models. 
For ease of description, we denote the spatiotemporal feature extractor as $g_{st}$, and the output spatiotemporal feature as $\mathbf{X}_{st}$, illustrated as,
\begin{equation}
\begin{aligned}
    \mathbf{X}_{st} &= g_{st}(\mathcal{X}) \\
     &= T (G(\mathcal{X};\mathbf{W}_G), \mathbf{W}_T)
\end{aligned}
\label{extractor}
\end{equation}
where we use $T$ to represent the function capturing temporal correlations, $G$ to represent the function capturing spatial correlations, and $\mathbf{W}_T$ and $\mathbf{W}_G$ denote learnable parameters. Equation (\ref{extractor}) signifies that the feature extractor mines temporal and spatial correlations and generates spatiotemporal features.

\subsection{Fairness-aware spatiotemporal architecture}
With the backbone of spatiotemporal learning well-trained, we introduce the our fairness-aware spatiotemporal learning architecture. Current machine learning solutions to fairness usually disentangle sensitive factors and derive a learning objective to learn the representations independent of all sensitive factors~\cite{qi2022profairrec, wang2022improving, wu2021fairness}. However, given the tasks of spatiotemporal learning, they usually lack explicit sensitive attributes, increasing the difficulty to identify which samples are prone to suffering unfair treatment. Therefore, from the representation perspective, we argue that fairness-aware spatiotemporal learning can be decomposed into two aspects. First, it is critical to explicitly identify  which samples are easy-to-learn and otherwise, which are difficult-to-learn. Second, how to sufficiently exploit the advantages of well-learned representations to improve the quality of samples which are prone to unfair treatment.  To this end, our fairness-aware spatiotemporal architecture is composed of two components, the fairness recognizer and the collaborative feature enhancement to systematically address above issues.

% Last but not least, following existing fairness-aware learning~\cite{liu2023generalized, he2023learning}, imposing a fairness-aware objective on the sample-level, to force the prediction performance of all  samples converge to the similar patterns.

%acquires to actively identify which samples are easy-to-learn and otherwise, which are difficult-to-learn. Correspondingly, to facilitate the methodology description, we now formally define the 'easy samples' and 'challenging samples'.
\subsubsection{Fairness recognizer}
To remedy the lacking sensitive attributes in  spatiotemporal datasets, we propose a fairness recognizer to identify whether the samples are difficult-to-learn. We first formally introduce the definition of easy samples and challenging samples.

% To mitigate the disparity of prediction performance, the core idea of collaborative fairness-aware learning is to discover  the samples with high-quality spatiotemporal representation, and then fully exploit the information of well-learned ones to enhance the representation of challenging ones. At the very first, we  formally introduce the easy samples and challenging samples.

% \noindent \textbf{Definition 3: Easy samples and challenging samples.}
\begin{definition}[Easy samples and challenging samples]
We characterize the learning difficulty of different spatiotemporal samples through the model's fitting degree of each sample. Given a sample $\bm{x}_i$,  we utilize the  error between the prediction and ground truth to quantify the model's fitting degree. Specifically, samples with $K$-smallest  errors  are categorized as the easy samples, denoted as $\mathcal{S}_e$, while other samples are categorized as the challenging samples, denoted as $\mathcal{S}_c$\footnote{Based on Pareto principle~\cite{Newman2004PowerLP}, we set $K$ to 20\%.}.
\end{definition}
%while samples with $K$-largest errors are categorized as the challenging samples, denoted as $\mathcal{S}_c$\footnote{Here, we set $K$ to 30\%.}.
%$\mathcal{S}_e = \{\bm{X}_j| j \in {\rm{arg min}}|{\widehat{\bm{Y}}}_j-\bm{Y}_j| \}$
%$\mathcal{S}_c = \{\bm{X}_j| j \in {\rm{arg max}}|{\widehat{\bm{Y}}}_j-\bm{Y}_j| \}$
%Then the stages of our collaborative feature enhancement can be summarized as challenging sample identification and representation-level mutual enhancement.  We will further dissect the techniques, a fairness recognizer and  collaborative feature enhancement,  to achieve the above two goals respectively. 

To proactively identify the samples into easy ones and challenging ones, we propose a learnable \textbf{fairness recognizer}, which is motivated by the concept of computational identifiability~\cite{Hébert-Johnson2018MulticalibrationCF, Lahoti2020FairnessWD}.
Given a family of binary functions $\mathcal{F}$, it is  said that a subgroup $\mathcal{S}$ is computationally-identifiable if there exists a function $f:\mathbf{X} \rightarrow \lbrace 0,1\rbrace$ in $\mathcal{F}$ such that $f(\bm{x})=1$ if and only if $f(\bm{x}) \in \mathcal{S}$. Building on this definition, we propose a learnable fairness recognizer to identify the computationally-identifiable regions with relative high  errors.  Our learning-based recognizer establishes the connections between spatiotemporal samples and the identification of easy or challenging judgement. We therefore devise a self-supervised task to guide the learning difficulty of samples and generate the \textbf{fairness signals}, where the fairness signals indicate whether the given sample is   challenging to learn, correspondingly to whether they are prone to be treated unfairly.

In the training phase, we  obtain the prediction error of each sample where we can rank the samples by MAE errors and make a partition  based on the rankings where easy samples with lower errors and challenging ones with higher errors.  Then we can take  these partition results as self-supervised signals and construct a self-supervised task via formulating a  binary classification. We assign  $z=1$ as  the self-supervised signal for easy samples, while assigning  the labels for challenging samples as $z=0$.

After then, we  are going to determine the inputs of learnable  fairness recognizer. First, recent literature has revealed that the spatiotemporal heterogeneity is mostly associated with the external factors (such as sampling time, sampling locations, weekdays, weather, etc.) ~\cite{Sang2022GCMTAG, Zhou2022ForeseeUS},  we  thus introduce these external factors into fairness recognizer to learn finer-grained features. Secondly, since spatiotemporal samples are essentially time series, adding statistical information of time series (such as mean, variance, etc.) will bring in informative knowledge including sequence trends and data distributions, empowering our framework to better characterize the patterns of time series. To this end, the auxiliary input of the fairness recognizer can be summarized as two aspects, the external spatiotemporal factors and statistical information of time series. We concatenate all these auxiliary factors into a fixed-length vector $\mathcal{C}=\{\bm{c}_1, \bm{c}_2, \cdot \cdot \cdot, \bm{c}_M\}$ as the inputs of our fairness recognizer, where $\bm{c}_i = [\bm{x}_{st}^i; \bm{e}_i; \mu_i; \sigma^2_i] \in \mathbb{R}^{d_c}$, where $\bm{x}_{st}^i$, $\bm{e}_i$, $\mu_i$ and $\sigma^2_i$ respectively represent the spatiotemporal features, external factor embedding, sequence mean and sequence variance of a given spatiotemporal sample $\bm{x}_i$. We will elaborate on how to process external factors in the section of experiments.

\textbf{Architecture design of the fairness recognizer.} 
%The fairness recognizer is responsible for mining the learning difficulty of different samples and identifying which samples are prone to suffering unfair treatment.
In fact, the learning difficulty is also concerned with spatiotemporal patterns, we thus exploit the Graph Convolutional Network (GCN) blocks as the basic architecture to instantiate our fairness recognizer. To be specific, we do not force our ST Graphs a predefined adjacency matrix, as the predefined topology may lack a direct relationship with the task, leading to significant bias. Inspired by previous study ~\cite{wu2020connecting}, we adopt an adaptive topology learning method to capture uni-directional relationships and generate the adjacency matrix as follows,
\begin{equation}
    \mathbf{D}^{-\frac{1}{2}} \mathbf{A} \mathbf{D}^{-\frac{1}{2}} = \text{ReLU}(\text{tanh}(\mathbf{E}_1 \mathbf{E}_2^\top - \mathbf{E}_2 \mathbf{E}_1^\top))
\end{equation}
where $\mathbf{E}_1$, $\mathbf{E}_2$ represent randomly initialized node embedding. To reduce repeated and redundant calculations during the iterative training process, we directly produce $\mathbf{D} ^{-\frac{1}{2}}  \mathbf{A} \mathbf{D}^{-\frac{1}{2}}$ rather than computing $\mathbf{A}$ and  its Laplacian matrix. Thus, the $l$-th GCN layer with adaptive topology learning can be formulated as,
\begin{equation}
    \mathbf{H}^{(l)} = \text{ReLU}(\mathbf{I} + \text{ReLU}(\text{tanh}( \mathbf{E}_1 \mathbf{E}_2^\top- \mathbf{E}_2 \mathbf{E}_1^\top))) \mathbf{H}^{(l-1)} \mathbf{W}^{l}
\end{equation}
where $\mathbf{H}^{(l)}$ represents the hidden states in the $l$-th GCN layer, $\mathbf{W}^l$ represents the parameter in the $l$-th GCN layer, where in our work, the 
fairness recognizer consists of three stacked   GCN layers. By denoting the fairness recognizer as $g_{fa}$, the output fairness signal  of a given spatiotemporal sample $\bm{x}_i$ can be formulated as,
\begin{equation}
    \hat{\bm{z}_i} = g_{fa}(\bm{c}_i|\mathcal{G})
\end{equation}

It is worth noting that the architecture design of the fairness recognizer plays an important role in controlling the granularity of computationally-identifiable regions. A more expressive recognizer architecture leads to a finer-grained identification but suffers more risks of overfitting to outliers. We will further analyze the impact of different network architecture designs  of the fairness recognizer in the section of  experiments.

\subsubsection{Collaborative feature enhancement}
After predicting the learning difficulty of spatiotemporal samples, the next challenge is how to proactively compensate for the perceived challenging samples and improving the model's performance on this subgroup. We posit that spatiotemporal samples with similar patterns also exhibit similar representations. Based on this observation, for challenging samples, we obtain compensatory samples with similar patterns but better learning quality and design an attention-based strategy for advantageous feature transfer and fusion, which achieves adaptive transfer and enhancement of advantageous representations between easy set and challenging set.

\textbf{Compensatory sample generation.} Spatiotemporal samples with similar observations often exhibit similar patterns in representation spaces. To capture such relationships, we establish a similarity matrix $\mathbf{S}$ between sample pairs as follows,
\begin{equation}
    \mathbf{S}_{ij}= \begin{cases} \text{SIM}(\bm{x}_{st}^i, \bm{x}_{st}^j) & {\text{for } i \neq j} \ \text{and} \ \bm{x}_j \in \mathcal{S}_e \\
0 & {\text{others}}
\end{cases}
\end{equation}
where $\bm{x}_{st}^i$ denotes the spatiotemporal feature of sample $\bm{x}_i$, and $\text{SIM}$ represents the similarity measurement and we instantiate that as the cosine similarity in our work. 
We select the top-$k_c$ samples both well-learned and most similar to it by learning in a  collaborative manner, where the selected well-learned samples are designated as compensatory samples. For a given challenging sample $\bm{x}_i$, we select top-$k_c$ most similar samples as its compensatory samples, denoted as $\{\bm{u}^{(i,1)}, \cdot \cdot \cdot, \bm{u}^{(i,k)}\}$. Then we aggregate the compensatory samples and denote the compensatory representation for $\bm{x}_i$ as $\bm{u}_{st}^i$ as,
\begin{equation}
    \bm{u}_{st}^i = \text{AGGREGATE}(\{\bm{u}_{st}^{(i,1)},\cdot \cdot \cdot, \bm{u}_{st}^{(i,k)}\})
\end{equation}
where $\bm{u}_{st}^{(i,j)}$ represents the spatiotemporal feature of compensatory sample $\bm{u}^{(i,j)}$. There are many options for ${\rm AGGREGATE}$ function, and  we instantiate it as MEAN-POOLING.

% we can sample $k$ compensatory samples $\{u_^{(1)}, ..., u^{(k)}\}$ from the multimodal distribution $\text{Multi}(s_i)$, where $s_i$ is the row vector of $S$ corresponding to $x_{st}^i$.  Then we aggregate the sample set $\{u_{st}^{(1)}, ..., u_{st}^{(k)}\}$, and denote the compensatory representations generated for $x_{st}^i$ as $u_{st}^i$, illustrated as,
% \begin{equation}
%     u_{st}^i = AGGREGATE(\{u_{st}^{(1)},...u_{st}^{(k)}\})
% \end{equation}
% There are many options for ${\rm AGGREGATE}$ function, and in our work we select the MEAN-POOLING method.

\textbf{Adaptive mix-up for fair representation.} We then achieve the awareness of the learning difficulty of each sample and generate compensatory representation for each challenging sample. Due to the spatiotemporal heterogeneity among different samples, we must devise a personalized fusion method to allow the adaptive representation aggregation. Therefore, we propose an attention-based mix-up strategy. Formally, we  can derive the mix-up strategy as follows,
\begin{equation}
    \begin{aligned}
\mathbf{Q} &= \bm{x}_{st}^i \mathbf{W}_q \\
\mathbf{K} &= \bm{u}_{st}^i \mathbf{W}_k \\
\alpha &= \text{softmax}(\frac{\mathbf{Q} \mathbf{K}^\top}{\sqrt{d_k}}) \\
\alpha' &= \text{MLP}(\alpha \bm{u}_{st}^i)
    \end{aligned}
\end{equation}
where $\bm{x}_{st}^i \in \mathbb{R}^d$, $\bm{u}_{st}^i \in \mathbb{R}^d$, $\mathbf{W}_q \in \mathbb{R}^{d \times d_k}$ and $\mathbf{W}_k\in \mathbb{R}^{d \times d_k}$ are learnable parameters, and we constrain $\alpha'$ within the range $(0,0.5)$ to preserve the intrinsic representation.

Finally, for any challenging sample $\bm{x}_i$, we obtain the mix-up representation $\bm{x}_{com}^i$ by fusing the intrinsic representation $\bm{x}_{st}^i$ and the compensatory representation $\bm{u}_{st}^i$ by,
\begin{equation}
    \bm{x}^i_{com} = (1-\alpha')\bm{x}_{st}^i + \alpha \bm{u}_{st}^i
\end{equation}
$\bm{x}_{com}^i$ is the well-remedied representation generated with sample-level joint optimization and knowledge transfer.

In summary, at representation level, the fairness recognizer and  the collaborative feature enhancement can work cooperatively, initially employing the concept of computationally-identifiable subgroup to design an auxiliary self-supervised task for perceiving the learning difficulty of diverse spatiotemporal samples. Thus our FairSTG can overcome lacking sensitive attributes. For challenging samples  which are prone to suffering unfairness, samples with similar patterns but well-learned representation are selected for compensation, enhancing the overall performance.

\subsection{Fairness-aware learning objective}
\label{sec:fairness-obj}
Recent literature has demonstrated that incorporating fairness metrics into learning objective can effectively remove discrimination during the training process~\cite{d2017conscientious}. To this end, in addition to collaboratively enhancing representations  of  challenging samples, our FairSTG imposes fairness constraints to improve the learning objective, which forces the model to provide fair treatment for diverse samples. After obtaining the compensated spatiotemporal representations, the output module produces the final predictions at one forward step, and the fairness-aware optimization objective constrains the model to optimize towards treating all samples fairly. Besides, considering the quality of  fairness recognizer is highly relied on the feature extractor, we thus further devise a two-stage training strategy with a warm-up phase to pre-generate the extractor parameters. Therefore, in this subsection, we first introduce our output module and then formally present our optimization objective, and finally provide the two-stage training strategy, serving for fairness-aware learning.

\subsubsection{Output module} The output module is instantiated with  two $1 \times 1$ standard convolutional layers, which transforms the input channel dimension to the  output dimension in an one-step-forward manner rather than a step-by-step style. For a spatiotemporal sample $\bm{x}_i \in \mathbb{R}^w$, the resulting output is $\hat{\bm{y}_i} \in \mathbb{R}^h$, where $h$ is the fixed dimension for outputs. 

\subsubsection{Optimization objective} 
Given the fairness-aware learning architecture, we can systematically derive the objective of our FairSTG, which can be summarized as three aspects, i.e., reweighted  loss for main spatiotemporal task by emphasizing the challenging samples, sample-level fairness-aware loss, and self-supervised objective for learning a fairness recognizer. 

%We focus on the task of spatiotemporal forecasting, designing a cost-sensitive regression loss and a fairness loss to improve the training process. Besides, we incorporate the binary classification loss for self-supervised learning. All the loss functions are organized into the form of a multi-task loss function.

\textbf{Reweighted loss for main task.} Actually, from the perspective of learning difficulty, each sample does not share the same importance in the same training batch towards final optimizations, e.g., some samples are easy to learn while others suffer higher errors. We thus introduce a reweighted loss  to regularize the main task. We first present the initial main task of spatiotemporal learning, which is instantiated as a regression problem associated with MAE, 
\begin{equation}
    \text{MAE}(\mathcal{Y}, \hat{\mathcal{Y}}) = \frac{1}{M}\sum_{i=1}^{M}|\hat{\bm{y}_i}-\bm{y}_i|
\end{equation}
To increase the emphasis of difficult-to-learn samples, we assign higher weights to samples with higher errors, which guides the model to pay more attention on these subgroups. The weights are normalized within batches to prevent gradient explosions. Specifically, we perform a normalization step that rescales the error to $[0,1]$ and produce the cost-sensitive weight $\lambda_i$, and we center the weight and add $1$ to ensure that all training samples to make sense to final loss. For any spatiotemporal sample $\bm{x}_i$, the corresponding weight $\lambda_i$ can be formulated as follows,
\begin{equation}
    \lambda_i = 1 + \frac{|\hat{\bm{y}_i}-\bm{y}_i|}{\sum_{j=1}^M{|\hat{\bm{y}_j}-\bm{y}_j|}}
\end{equation}
Furthermore, we construct the reweighted  loss  based on the cost-sensitive weights as,
\begin{equation}
    \mathcal{L}_r = \frac{1}{M}\sum_{i=1}^{M} \lambda_i \cdot |\hat{\bm{y}_i}-\bm{y}_i|
\end{equation}
This reweighted loss can enable the learning objective more sensitive to challenging samples, thus increasing the attention for modeling challenging ones. With the process of model training, the $\lambda$  will be dynamic with learning process and adjust the learning in fine-grained manner.

\textbf{Fairness loss.} Second, to explicitly ensure the fairness of prediction performance across different samples, we exploit the variance of MAE as the fairness constraint term to directly guide the model's learning process, i.e.,
\begin{equation}
    \mathcal{L}_f = \frac{1}{M}\sum_{i=0}^{M-1}[|\hat{\bm{y}_i}-\bm{y}_i|-\frac{1}{M}\sum_{j=0}^{M-1}|{\hat{\bm{y}_j}-\bm{y}_j}|]^2
\end{equation}
With this fairness loss, we actually introduce the awareness of fairness into our training process from the perspective of immediate strategy, which guarantees the equitable performance across samples. 

\textbf{Self-supervised loss.} Third,  the  self-supervised learning objective of our fairness recognizer is instantiated with the  Balanced Cross-Entropy loss (BCE Loss). To be specific, we take the ranking of prediction performance as the self-supervised signals by assigning  $z=1$ and  $z=0$ for easy and challenging samples respectively. Then the  fairness recognizer output $\hat{z}$ can be compared with self-supervised signal $z$ with the following contrastive loss, i.e.,
\begin{equation}
    \mathcal{L}_s(\hat{\bm{z}}, \bm{z}) = -\omega*(\bm{z}* \ln{\hat{\bm{z}}}) + (1-\bm{z})*\ln(1-\hat{\bm{z}})
\end{equation}
where $\omega$ represents the weight for positive samples, which is set based on the proportion of easy and challenging samples set. In our work, we set $\omega=4$ via experimental trials.

\textbf{Overall fairness-aware learning loss.} By summarizing the above  loss terms, the final loss function can be written as,
\begin{equation}
    \mathcal{L} = \mu_r\mathcal{L}_r + \mu_f\mathcal{L}_f + \mu_s\mathcal{L}_s
\end{equation}
where $\mu_r$, $\mu_f$ and $\mu_s$ are hyper-parameters.

It is worth highlighting that we introduce fairness constraints at the level of optimization objectives. On the one hand, we design a reweighted  loss  for the main task based on cost-sensitive weight, to emphasize more attention on challenging samples which are  prone to unfair treatment and make timely adjustment. On the other hand, we devise a fairness loss term to directly minimize the performance difference across samples during the whole learning pipeline. These two objectives can work cooperatively to improve prediction fairness technically, not only making necessary adjustment to sample weights, but also minimizing  the performance heterogeneity with an explicit objective constraint. 

\subsubsection{Training strategy}
Given that the accuracy of fairness recognizer is highly relied on  the quality of the representations generated by the spatiotemporal feature extractor, we design a two-stage training strategy with a  warm-up phase and a fairness-aware learning phase.
\begin{itemize}
\item Warm-up phase. We only train the spatiotemporal feature extractor and output module. In this phase, we set $\mu_r \neq 0$, $\mu_f=0$, $\mu_s=0$, and $\lambda_i=1$ for all samples.
\item Fairness-aware learning. All modules are trained and  we set $\mu_r \neq 0$, $\mu_f \neq0$ and $\mu_s \neq 0$. We will discuss the selection of hyper-parameters in detail in the experiment section.
% \item Inference phase. We select the MAE as the loss function for the main task and $\mathcal{L}_f$ as the fairness metric.
\end{itemize}
With such two-stage training strategy, we can formulate a well-learned fairness recognizer to produce accurate fairness signals and then the representations can be collaboratively enhanced with our fairness-aware learning architecture.  

\section{Experiments}
%In this section, we evaluate the effectiveness of our FairSTG by comparing both prediction performance and fairness metric. The ablation studies and case studies for fairness.
\subsection{Dataset}

We select four real-world mobility-related spatiotemporal datasets, from human mobility, air quality to smart grids, to evaluate the effectiveness and generality of our FairSTG. We summarize statistics of benchmark datasets in Table~\ref{tab:dataset}, where the samples refer to the input-output pair for the model. 

\begin{enumerate}
    \item[$\bullet$] METR-LA~\cite{wu2020connecting}: It indicates the traffics and inter-city human mobility, containing the average traffic speed measured by 207 loop sensors on the highways of Los Angeles County, from Mar 2012 to Jun 2012.
    \item[$\bullet$] PEMS-BAY~\cite{wu2020connecting}: It is the mobility dataset,  composed of average traffic speed measured by 325 sensors in the Bay Area, ranging from Jan 2017 to May 2017.
    \item[$\bullet$] KnowAir~\cite{wang2020pm2}:  This dataset records the PM2.5 concentrations, one of the important air quality index, in 184 main cities in China, where it is collected every 3 hours from Sep 2016  to Jan 2017.
    \item[$\bullet$] ETT~\cite{zhou2021informer}:  The ETT (Electricity Transformer Temperature) is a crucial indicator of electric power in cities. The dataset is collected every 1 hour from July 2016 to July 2018 where it consists of 7 features, and each consumer is considered as a distinct node.
\end{enumerate}

\begin{table}[]
\caption{The statistics of datasets}
\label{tab:dataset}
\resizebox{0.45\textwidth}{!}{
    \begin{tabular}{
    >{\columncolor[HTML]{FFFFFF}}l |
    >{\columncolor[HTML]{FFFFFF}}l |
    >{\columncolor[HTML]{FFFFFF}}l |
    >{\columncolor[HTML]{FFFFFF}}l }
    \hline
    {\color[HTML]{333333} Dataset}  &  \tabincell{l}{Number of \\ samples}  & \tabincell{l}{Number of \\ nodes}  & \tabincell{l}{Sampling \\ interval}  \\ \hline
    {\color[HTML]{333333} Metr-LA}  & {\color[HTML]{333333} 34272}   & {\color[HTML]{333333} 207}   & {\color[HTML]{333333} 5 minutes}       \\
    {\color[HTML]{333333} PEMS-BAY} & {\color[HTML]{333333} 52116}   & {\color[HTML]{333333} 325}   & {\color[HTML]{333333} 5 minutes}       \\
    {\color[HTML]{333333} KnowAir}  & {\color[HTML]{333333} 11688}   & {\color[HTML]{333333} 184}   & {\color[HTML]{333333} 3 hours}         \\
    {\color[HTML]{333333} ETT}      & {\color[HTML]{333333} 17420}   & {\color[HTML]{333333} 7}     & {\color[HTML]{333333} 1 hour}          \\ \hline
    \end{tabular}
}
\end{table}

\subsection{Implementation details and  evaluation metrics}
\textbf{Implementation details.} Following  practices of classical time-series data split~\cite{wu2020connecting,li2017diffusion}, we split all the available data into training, validation, and testing parts with the ratio of 7:2:1. We normalize the data with standard normalization  to ensure the stability of training process. We set the input window $w=12$ and the output window $h=12$. The model is trained by the Adam optimizer with gradient clip $5$. The learning rate is set as 0.001. In our experiments, the dimension of the spatiotemporal representations and compensatory representations is fixed to $64$. We examine the number of compensatory samples $k_c$ in range of $\{5,10,20\}$ and choose 5 for all datasets. Regarding optimization objective, the parameter $\mu_r$ is fixed to 1 while $\mu_s$ is fixed to 0.1, and we examine the trade-off fairness parameter $\mu_f$ in range of $\{0.01, 0.1, 0.5, 1.0, 1.5\}$ for different datasets, we finally set 0.5 for METR-LA and PEMS-BAY, 0.1 for KnowAir and ETT. 

Regarding some baselines requiring the predefined adjacent matrix, we extract the pairwise distances to measure the node-wise proximity as the adjacencies. For METR-LA and PEMS-BAY, following previous work~\cite{li2017diffusion}, we compute  distances between sensors within the road network and build the adjacency matrix using thresholded Gaussian kernel. For KnowAir, we compute the geographical distance between sampling points to create an adjacency matrix, and retain the top 20\% of nodes with the highest weights as neighbors for each node to ensure the adjacency sparsity. For ETT, we take the most recent month to construct the adjacency with cosine similarity.

Concerning external factors, we separate them into two groups, i.e., continuous and categorical features. Continuous factors including temperature and time stamps are directly concatenated into a vector $\bm{e}^{con}$. Categorical factors including weekday, weather conditions are separately projected into a low-dimensional continuous vector space through embedding layers. These embeddings are then concatenated to form the vector $\bm{e}^{cat}$. Specifically, to enable the fairness recognizer to learn the learning difficulty more accurately, we combine not only spatiotemporal external factors but also the sequence statistical information. Without loss of generality, for each sample $\bm{x}_i$ and the corresponding spatiotemporal representation $\bm{x}_{st}^i$, we compute the  statistical information, mean $\mu_i$ and variance $\sigma_i^2$ of corresponding sample $\bm{x}_i$ in the element level. And we convert the time stamps of day into continuous values as $\bm{e}^{con}$. For indicator of day of week and node indexes, we  organize them as  categorical vector $\bm{e}^{cat}$. And continuous embedding $\bm{e}^{con}$ and categorical embedding $\bm{e}^{cat}$ are concatenated to form the external embedding $\bm{e_i}$ for the corresponding sample. In addition,  those vectors  are concatenated as the input of our fairness recognizer, denoted as $\mathcal{C}=\{\bm{c}_1, \bm{c}_2, \cdot \cdot \cdot, \bm{c}_M\}$, where $\bm{c}_i = [\bm{x}_{st}^i; \bm{e}_i; \mu_i; \sigma^2_i] \in \mathbb{R}^{d_c}$.
% These statistical information and immediate time indicators are concatenated as $\bm{e}^{con}$. 

\textbf{Evaluation metrics. } The goal  of FairSTG is to improve the sample-level prediction fairness with simultaneously retaining the overall performances, and we also argue that the error-based metrics without considering prediction heterogeneity deduce the evaluation function in a holistic aspect. To this end, we incorporate five evaluation metrics that can be divided into two aspects, prediction accuracy evaluation and quality of fairness learning, where the fairness quality evaluation helps remedy the function degeneration of only error metrics.

We take Mean Absolute Error (MAE), Mean Absolute Percentage Error (MAPE), and Root Mean Squared Error (RMSE) as evaluation metrics for accuracy, while introduce  the variance of sample-level errors over the  testing set, i.e., sample-level MAE-var and MAPE-var,  as the fairness metrics. %For all five metrics,  the lower, the better.

\subsection{Baselines}
We compare our proposed framework with 6 state-of-the-art baselines as follows.
\begin{enumerate}
    \item[$\bullet$] DCRNN~\cite{li2017diffusion}: A graph-based recurrent neural network, which combines graph diffusion convolutions with recurrent neural network.
    \item[$\bullet$] STGCN~\cite{yu2018spatio}: A spatiotemporal graph convolutional network, incorporating graph convolutions with 1D convolutions.
    \item[$\bullet$] AGCRN~\cite{bai2020adaptive}: A spaiotemporal graph convolutional network, which captures node-specific patterns and infers the inter-dependencies among time series automatically.
    \item[$\bullet$] MTGNN~\cite{wu2020connecting}: A spatiotemporal graph convolutional network, which integrates adaptive graph learning, graph convolutions and temporal convolutions.
    \item[$\bullet$] D$^2$STGNN~\cite{shao2022decoupled}: A decoupled dynamic spatiotemporal neural networks, which decouples and handles the diffusion and inherent information separately.
    \item[$\bullet$] ST-SSL~\cite{ji2023spatio}: A spaiotemporal learning traffic prediction framework, which designs two self-supervised auxiliary tasks with 
    a contrastive learning objective, to  gain awareness of spatial-temporal heterogeneity and  supplement the main forecasting task.
\end{enumerate}

\subsection{Experimental results}
Since our FairSTG is a general pluggable spatiotemporal learning framework alleviating the unfairness issue, we choose two advanced spatiotemporal learning models, MTGNN and D$^2$STGNN, as the backbones  for the spatiotemporal feature extractor. We implement 3-step, 6-step and 12-step prediction, which is illustrated as horizons in our result tables, and all the solutions are evaluated based on five metrics  regarding both fairness metrics and forecasting performance. Table ~\ref{tabel: main_res} elaborate the forecasting performance and fairness metrics across different models. The best results are \textbf{bolded} and the runner-up are \underline{underlined}.


\begin{table*}[]
\begin{center}
\caption{Main comparison results.}
\label{tabel: main_res}
\resizebox{\textwidth}{!}{
\begin{tabular}{lccccccccccccccc}
\hline
\rowcolor[HTML]{FFFFFF} 
\multicolumn{16}{c}{\cellcolor[HTML]{FFFFFF}{\color[HTML]{000000} METR-LA}}  \\ \hline
\rowcolor[HTML]{FFFFFF} 
\multicolumn{1}{l|}{\cellcolor[HTML]{FFFFFF}{\color[HTML]{333333} }}               & \multicolumn{5}{c|}{\cellcolor[HTML]{FFFFFF}{\color[HTML]{333333} horizon-3}}                           & \multicolumn{5}{c|}{\cellcolor[HTML]{FFFFFF}{\color[HTML]{333333} horizon-6}} & \multicolumn{5}{c}{\cellcolor[HTML]{FFFFFF}{\color[HTML]{333333} horizon-12}}  \\ \cline{2-16} 
\rowcolor[HTML]{FFFFFF} 
\multicolumn{1}{l|}{\cellcolor[HTML]{FFFFFF}{\color[HTML]{333333} }}               & {\color[HTML]{333333} MAE}                                                         & {\color[HTML]{333333} MAPE}            & {\color[HTML]{333333} RMSE}            & {\color[HTML]{333333} \begin{tabular}[c]{@{}c@{}}MAE\\ var\end{tabular}} & \multicolumn{1}{c|}{\cellcolor[HTML]{FFFFFF}{\color[HTML]{333333} \begin{tabular}[c]{@{}c@{}}MAPE\\ var\end{tabular}}} & {\color[HTML]{333333} MAE}             & {\color[HTML]{333333} MAPE}            & {\color[HTML]{333333} RMSE}            & {\color[HTML]{333333} \begin{tabular}[c]{@{}c@{}}MAE\\ var\end{tabular}} & \multicolumn{1}{c|}{\cellcolor[HTML]{FFFFFF}{\color[HTML]{333333} \begin{tabular}[c]{@{}c@{}}MAPE\\ var\end{tabular}}} & {\color[HTML]{333333} MAE}             & {\color[HTML]{333333} MAPE}            & {\color[HTML]{333333} RMSE}            & {\color[HTML]{333333} \begin{tabular}[c]{@{}c@{}}MAE\\ var\end{tabular}} & {\color[HTML]{333333} \begin{tabular}[c]{@{}c@{}}MAPE\\ var\end{tabular}} \\ \hline
\rowcolor[HTML]{FFFFFF} 
\multicolumn{1}{l|}{\cellcolor[HTML]{FFFFFF}{\color[HTML]{333333} DCRNN}}          & \multicolumn{1}{l}{\cellcolor[HTML]{FFFFFF}{\color[HTML]{333333} 2.7022}}          & {\color[HTML]{333333} 6.96\%}          & {\color[HTML]{333333} 5.2389}          & {\color[HTML]{333333} 23.8967}                                           & \multicolumn{1}{c|}{\cellcolor[HTML]{FFFFFF}{\color[HTML]{333333} 0.0755}}                                             & {\color[HTML]{333333} 3.1126}          & {\color[HTML]{333333} 8.51\%}          & {\color[HTML]{333333} 6.3414}          & {\color[HTML]{333333} 36.0244}                                           & \multicolumn{1}{c|}{\cellcolor[HTML]{FFFFFF}{\color[HTML]{333333} 0.1201}}                                             & {\color[HTML]{333333} 3.5705}          & {\color[HTML]{333333} 10.40\%}         & {\color[HTML]{333333} 7.5242}          & {\color[HTML]{333333} 51.6100}                                           & {\color[HTML]{333333} 0.1825}                                             \\
\rowcolor[HTML]{FFFFFF} 
\multicolumn{1}{l|}{\cellcolor[HTML]{FFFFFF}{\color[HTML]{333333} STGCN}}          & \multicolumn{1}{l}{\cellcolor[HTML]{FFFFFF}{\color[HTML]{333333} 3.4535}}          & {\color[HTML]{333333} 8.51\%}          & {\color[HTML]{333333} 7.7567}          & {\color[HTML]{333333} 56.4673}                                           & \multicolumn{1}{c|}{\cellcolor[HTML]{FFFFFF}{\color[HTML]{333333} 0.0757}}                                             & {\color[HTML]{333333} 4.4417}          & {\color[HTML]{333333} 11.45\%}         & {\color[HTML]{333333} 10.1028}         & {\color[HTML]{333333} 96.2961}                                           & \multicolumn{1}{c|}{\cellcolor[HTML]{FFFFFF}{\color[HTML]{333333} 0.1430}}                                             & {\color[HTML]{333333} 5.9071}          & {\color[HTML]{333333} 15.75\%}         & {\color[HTML]{333333} 12.9456}         & {\color[HTML]{333333} 155.6183}                                          & {\color[HTML]{333333} 0.2589}                                             \\
\rowcolor[HTML]{FFFFFF} 
\multicolumn{1}{l|}{\cellcolor[HTML]{FFFFFF}{\color[HTML]{333333} AGCRN}}          & \multicolumn{1}{l}{\cellcolor[HTML]{FFFFFF}{\color[HTML]{333333} 3.3413}}          & {\color[HTML]{333333} 8.33\%}          & {\color[HTML]{333333} 7.5762}          & {\color[HTML]{333333} 54.1650}                                           & \multicolumn{1}{c|}{\cellcolor[HTML]{FFFFFF}{\color[HTML]{333333} 0.0789}}                                             & {\color[HTML]{333333} 4.0211}          & {\color[HTML]{333333} 10.15\%}         & {\color[HTML]{333333} 9.3950}          & {\color[HTML]{333333} 84.2929}                                           & \multicolumn{1}{c|}{\cellcolor[HTML]{FFFFFF}{\color[HTML]{333333} 0.1240}}                                             & {\color[HTML]{333333} 4.9653}          & {\color[HTML]{333333} 12.53\%}         & {\color[HTML]{333333} 11.6030}         & {\color[HTML]{333333} 128.5844}                                          & {\color[HTML]{333333} 0.1820}                                             \\
\rowcolor[HTML]{FFFFFF} 
\multicolumn{1}{l|}{\cellcolor[HTML]{FFFFFF}{\color[HTML]{333333} MTGNN}}          & \multicolumn{1}{l}{\cellcolor[HTML]{FFFFFF}{\color[HTML]{333333} 2.6793}}          & {\color[HTML]{333333} 6.90\%}          & {\color[HTML]{333333} 5.1800}          & {\color[HTML]{333333} 22.6463}                                           & \multicolumn{1}{c|}{\cellcolor[HTML]{FFFFFF}{\color[HTML]{333333} 0.0738}}                                             & {\color[HTML]{333333} 3.0409}          & {\color[HTML]{333333} 8.19\%}          & {\color[HTML]{333333} 6.1705}          & {\color[HTML]{333333} 33.3740}                                           & \multicolumn{1}{c|}{\cellcolor[HTML]{FFFFFF}{\color[HTML]{333333} 0.1121}}                                             & {\color[HTML]{333333} 3.4952}          & {\color[HTML]{333333} 9.87\%}          & {\color[HTML]{333333} 7.2361}          & {\color[HTML]{333333} 47.6894}                                           & {\color[HTML]{333333} 0.1598}                                             \\
\rowcolor[HTML]{FFFFFF} 
\multicolumn{1}{l|}{\cellcolor[HTML]{FFFFFF}{\color[HTML]{333333} D$^2$STGNN}}        & \multicolumn{1}{l}{\cellcolor[HTML]{FFFFFF}{\color[HTML]{333333} \textbf{2.5602}}} & {\color[HTML]{333333} \textbf{6.36\%}} & {\color[HTML]{333333} {\ul 4.9837}}    & {\color[HTML]{333333} 23.1246}                                           & \multicolumn{1}{c|}{\cellcolor[HTML]{FFFFFF}{\color[HTML]{333333} {\ul 0.0524}}}                                       & {\color[HTML]{333333} \textbf{2.9194}} & {\color[HTML]{333333} \textbf{7.72\%}} & {\color[HTML]{333333} {\ul 6.0272}}    & {\color[HTML]{333333} 34.8865}                                           & \multicolumn{1}{c|}{\cellcolor[HTML]{FFFFFF}{\color[HTML]{333333} {\ul 0.0932}}}                                       & {\color[HTML]{333333} \textbf{3.3567}} & {\color[HTML]{333333} {\ul 9.43\%}}    & {\color[HTML]{333333} 7.1357}          & {\color[HTML]{333333} 49.4757}                                           & {\color[HTML]{333333} {\ul 0.1431}}                                       \\
\rowcolor[HTML]{FFFFFF} 
\multicolumn{1}{l|}{\cellcolor[HTML]{FFFFFF}{\color[HTML]{333333} ST-SSL}}         & \multicolumn{1}{l}{\cellcolor[HTML]{FFFFFF}{\color[HTML]{333333} 2.8243}}          & {\color[HTML]{333333} 7.39\%}          & {\color[HTML]{333333} 5.4742}          & {\color[HTML]{333333} 26.1182}                                           & \multicolumn{1}{c|}{\cellcolor[HTML]{FFFFFF}{\color[HTML]{333333} 0.0816}}                                             & {\color[HTML]{333333} 3.2722}          & {\color[HTML]{333333} 9.14\%}          & {\color[HTML]{333333} 6.6587}          & {\color[HTML]{333333} 39.7393}                                           & \multicolumn{1}{c|}{\cellcolor[HTML]{FFFFFF}{\color[HTML]{333333} 0.1382}}                                             & {\color[HTML]{333333} 3.7280}          & {\color[HTML]{333333} 10.67\%}         & {\color[HTML]{333333} 7.6999}          & {\color[HTML]{333333} 53.5600}                                           & {\color[HTML]{333333} 0.1781}                                             \\ \hline
\rowcolor[HTML]{FFFFFF} 
\multicolumn{1}{l|}{\cellcolor[HTML]{FFFFFF}{\color[HTML]{333333} FAirST+MTGNN}}   & \multicolumn{1}{l}{\cellcolor[HTML]{FFFFFF}{\color[HTML]{333333} 2.7316}}          & {\color[HTML]{333333} 7.02\%}          & {\color[HTML]{333333} 5.0509}          & {\color[HTML]{333333} \textbf{21.5644}}                                  & \multicolumn{1}{c|}{\cellcolor[HTML]{FFFFFF}{\color[HTML]{333333} 0.0684}}                                             & {\color[HTML]{333333} 3.1291}          & {\color[HTML]{333333} 8.39\%}          & {\color[HTML]{333333} 6.0534}          & {\color[HTML]{333333} \textbf{31.9005}}                                  & \multicolumn{1}{c|}{\cellcolor[HTML]{FFFFFF}{\color[HTML]{333333} 0.1037}}                                             & {\color[HTML]{333333} 3.5521}          & {\color[HTML]{333333} 10.17\%}         & {\color[HTML]{333333} {\ul 7.1207}}    & {\color[HTML]{333333} {\ul 45.5186}}                                     & {\color[HTML]{333333} 0.1525}                                             \\
\rowcolor[HTML]{FFFFFF} 
\multicolumn{1}{l|}{\cellcolor[HTML]{FFFFFF}{\color[HTML]{333333} FAirST+D$^2$STGNN}} & {\color[HTML]{333333} {\ul 2.6020}}                                                & {\color[HTML]{333333} {\ul 6.47\%}}    & {\color[HTML]{333333} \textbf{4.8756}} & {\color[HTML]{333333} {\ul 21.6349}}                                     & \multicolumn{1}{c|}{\cellcolor[HTML]{FFFFFF}{\color[HTML]{333333} \textbf{0.0499}}}                                    & {\color[HTML]{333333} {\ul 2.9602}}    & {\color[HTML]{333333} {\ul 7.80\%}}    & {\color[HTML]{333333} \textbf{5.8432}} & {\color[HTML]{333333} {\ul 32.0360}}                                     & \multicolumn{1}{c|}{\cellcolor[HTML]{FFFFFF}{\color[HTML]{333333} \textbf{0.0859}}}                                    & {\color[HTML]{333333} {\ul 3.4316}}    & {\color[HTML]{333333} \textbf{9.40\%}} & {\color[HTML]{333333} \textbf{6.8938}} & {\color[HTML]{333333} \textbf{44.9188}}                                  & {\color[HTML]{333333} \textbf{0.1254}}                                    \\ \hline
\rowcolor[HTML]{FFFFFF} 
\multicolumn{16}{c}{\cellcolor[HTML]{FFFFFF}PEMS-BAY}   \\ \hline
\rowcolor[HTML]{FFFFFF} 
\multicolumn{1}{l|}{\cellcolor[HTML]{FFFFFF}{\color[HTML]{333333} }}               & \multicolumn{5}{c|}{\cellcolor[HTML]{FFFFFF}{\color[HTML]{333333} horizon-3}}                                        & \multicolumn{5}{c|}{\cellcolor[HTML]{FFFFFF}{\color[HTML]{333333} horizon-6}}   & \multicolumn{5}{c}{\cellcolor[HTML]{FFFFFF}{\color[HTML]{333333} horizon-12}}   \\ \cline{2-16} 
\rowcolor[HTML]{FFFFFF} 
\multicolumn{1}{l|}{\cellcolor[HTML]{FFFFFF}{\color[HTML]{333333} }}               & {\color[HTML]{333333} MAE}                                                         & {\color[HTML]{333333} MAPE}            & {\color[HTML]{333333} RMSE}            & {\color[HTML]{333333} \begin{tabular}[c]{@{}c@{}}MAE\\ var\end{tabular}} & \multicolumn{1}{c|}{\cellcolor[HTML]{FFFFFF}{\color[HTML]{333333} \begin{tabular}[c]{@{}c@{}}MAPE\\ var\end{tabular}}} & {\color[HTML]{333333} MAE}             & {\color[HTML]{333333} MAPE}            & {\color[HTML]{333333} RMSE}            & {\color[HTML]{333333} \begin{tabular}[c]{@{}c@{}}MAE\\ var\end{tabular}} & \multicolumn{1}{c|}{\cellcolor[HTML]{FFFFFF}{\color[HTML]{333333} \begin{tabular}[c]{@{}c@{}}MAPE\\ var\end{tabular}}} & {\color[HTML]{333333} MAE}             & {\color[HTML]{333333} MAPE}            & {\color[HTML]{333333} RMSE}            & {\color[HTML]{333333} \begin{tabular}[c]{@{}c@{}}MAE\\ var\end{tabular}} & {\color[HTML]{333333} \begin{tabular}[c]{@{}c@{}}MAPE\\ var\end{tabular}} \\ \hline
\rowcolor[HTML]{FFFFFF} 
\multicolumn{1}{l|}{\cellcolor[HTML]{FFFFFF}{\color[HTML]{333333} DCRNN}}          & {\ul 1.3152}                                                                       & \textbf{2.75\%}                        & 2.7785                                 & 5.9908                                                                   & \multicolumn{1}{c|}{\cellcolor[HTML]{FFFFFF}0.0083}                                                                    & \textbf{1.6521}                        & {\ul 3.71\%}                           & 3.7743                                 & 11.5165                                                                  & \multicolumn{1}{c|}{\cellcolor[HTML]{FFFFFF}0.0229}                                                                    & {\ul 1.9634}                           & {\ul 4.64\%}                           & 4.5921                                 & 17.2332                                                                  & 0.0579                                                                    \\
\rowcolor[HTML]{FFFFFF} 
\multicolumn{1}{l|}{\cellcolor[HTML]{FFFFFF}{\color[HTML]{333333} STGCN}}          & 1.3731                                                                             & 2.93\%                                 & 2.8451                                 & 6.2096                                                                   & \multicolumn{1}{c|}{\cellcolor[HTML]{FFFFFF}0.0128}                                                                    & 1.8013                                 & 4.16\%                                 & 4.0325                                 & 13.0175                                                                  & \multicolumn{1}{c|}{\cellcolor[HTML]{FFFFFF}0.0409}                                                                    & 2.3241                                 & 5.49\%                                 & 5.1715                                 & 21.3443                                                                  & 0.0642                                                                    \\
\rowcolor[HTML]{FFFFFF} 
\multicolumn{1}{l|}{\cellcolor[HTML]{FFFFFF}{\color[HTML]{333333} AGCRN}}          & 1.4135                                                                             & 3.09\%                                 & 3.0233                                 & 7.1421                                                                   & \multicolumn{1}{c|}{\cellcolor[HTML]{FFFFFF}0.0132}                                                                    & 1.7500                                 & 4.05\%                                 & 3.9967                                 & 12.9109                                                                  & \multicolumn{1}{c|}{\cellcolor[HTML]{FFFFFF}0.0311}                                                                    & 2.0721                                 & 4.91\%                                 & 4.7852                                 & 18.6050                                                                  & 0.0521                                                                    \\
\rowcolor[HTML]{FFFFFF} 
\multicolumn{1}{l|}{\cellcolor[HTML]{FFFFFF}{\color[HTML]{333333} MTGNN}}          & 1.3361                                                                             & {\ul 2.80\%}                           & 2.8079                                 & 6.0993                                                                   & \multicolumn{1}{c|}{\cellcolor[HTML]{FFFFFF}{\ul 0.0082}}                                                              & 1.6687                                 & 3.80\%                                 & 3.7648                                 & 11.3900                                                                  & \multicolumn{1}{c|}{\cellcolor[HTML]{FFFFFF}{\ul 0.0249}}                                                              & 1.9899                                 & 4.77\%                                 & 4.5535                                 & 16.7757                                                                  & {\ul 0.0508}                                                              \\
\rowcolor[HTML]{FFFFFF} 
\multicolumn{1}{l|}{\cellcolor[HTML]{FFFFFF}{\color[HTML]{333333} D$^2$STGNN}}        & \textbf{1.3093}                                                                    & 2.83\%                                 & \textbf{2.7424}                        & 5.8562                                                                   & \multicolumn{1}{c|}{\cellcolor[HTML]{FFFFFF}0.0089}                                                                    & {\ul 1.6540}                           & 3.92\%                                 & 3.7751                                 & 11.3713                                                                  & \multicolumn{1}{c|}{\cellcolor[HTML]{FFFFFF}0.0343}                                                                    & 1.9818                                 & 4.84\%                                 & 4.5864                                 & 17.0290                                                                  & 0.0624                                                                    \\
\rowcolor[HTML]{FFFFFF} 
\multicolumn{1}{l|}{\cellcolor[HTML]{FFFFFF}{\color[HTML]{333333} ST-SSL}}         & 1.4030                                                                             & 2.91\%                                 & 3.0104                                 & 7.0943                                                                   & \multicolumn{1}{c|}{\cellcolor[HTML]{FFFFFF}0.0081}                                                                    & 1.8160                                 & 4.16\%                                 & 4.0912                                 & 13.4398                                                                  & \multicolumn{1}{c|}{\cellcolor[HTML]{FFFFFF}0.0313}                                                                    & 2.1250                                 & 5.17\%                                 & 4.7808                                 & 18.3406                                                                  & 0.0640                                                                    \\ \hline
\rowcolor[HTML]{FFFFFF} 
\multicolumn{1}{l|}{\cellcolor[HTML]{FFFFFF}{\color[HTML]{333333} FAirST+MTGNN}}   & 1.3312                                                                             & 2.81\%                                 & {\ul 2.7445}                           & \textbf{5.7608}                                                          & \multicolumn{1}{c|}{\cellcolor[HTML]{FFFFFF}\textbf{0.0078}}                                                           & 1.6550                                 & \textbf{3.70\%}                        & \textbf{3.6613}                        & \textbf{10.6667}                                                         & \multicolumn{1}{c|}{\cellcolor[HTML]{FFFFFF}\textbf{0.0228}}                                                           & \textbf{1.9628}                        & \textbf{4.58\%}                        & \textbf{4.3885}                        & \textbf{15.4070}                                                         & \textbf{0.0448}                                                           \\
\rowcolor[HTML]{FFFFFF} 
\multicolumn{1}{l|}{\cellcolor[HTML]{FFFFFF}{\color[HTML]{333333} FAirST+D$^2$STGNN}} & 1.3678                                                                             & 2.96\%                                 & 2.7773                                 & {\ul 5.8435}                                                             & \multicolumn{1}{c|}{\cellcolor[HTML]{FFFFFF}0.0086}                                                                    & 1.6977                                 & 4.02\%                                 & {\ul 3.7382}                           & {\ul 11.0251}                                                            & \multicolumn{1}{c|}{\cellcolor[HTML]{FFFFFF}0.0331}                                                                    & 2.0149                                 & 4.97\%                                 & {\ul 4.4479}                           & {\ul 15.6454}                                                         & 0.0589                                                                    \\ \hline
\rowcolor[HTML]{FFFFFF} 
\multicolumn{16}{c}{\cellcolor[HTML]{FFFFFF}KnowAir}    \\ \hline
\rowcolor[HTML]{FFFFFF} 
\multicolumn{1}{l|}{\cellcolor[HTML]{FFFFFF}{\color[HTML]{333333} }}               & \multicolumn{5}{c|}{\cellcolor[HTML]{FFFFFF}{\color[HTML]{333333} horizon-3}}                                                                                                                                                                                                                                                                                            & \multicolumn{5}{c|}{\cellcolor[HTML]{FFFFFF}{\color[HTML]{333333} horizon-6}}                                                                                                                                                                                                                                                & \multicolumn{5}{c}{\cellcolor[HTML]{FFFFFF}{\color[HTML]{333333} horizon-12}}  \\ \cline{2-16} 
\rowcolor[HTML]{FFFFFF} 
\multicolumn{1}{l|}{\cellcolor[HTML]{FFFFFF}{\color[HTML]{333333} }}               & {\color[HTML]{333333} MAE}                                                         & {\color[HTML]{333333} MAPE}            & {\color[HTML]{333333} RMSE}            & {\color[HTML]{333333} \begin{tabular}[c]{@{}c@{}}MAE\\ var\end{tabular}} & \multicolumn{1}{c|}{\cellcolor[HTML]{FFFFFF}{\color[HTML]{333333} \begin{tabular}[c]{@{}c@{}}MAPE\\ var\end{tabular}}} & {\color[HTML]{333333} MAE}             & {\color[HTML]{333333} MAPE}            & {\color[HTML]{333333} RMSE}            & {\color[HTML]{333333} \begin{tabular}[c]{@{}c@{}}MAE\\ var\end{tabular}} & \multicolumn{1}{c|}{\cellcolor[HTML]{FFFFFF}{\color[HTML]{333333} \begin{tabular}[c]{@{}c@{}}MAPE\\ var\end{tabular}}} & {\color[HTML]{333333} MAE}             & {\color[HTML]{333333} MAPE}            & {\color[HTML]{333333} RMSE}            & {\color[HTML]{333333} \begin{tabular}[c]{@{}c@{}}MAE\\ var\end{tabular}} & {\color[HTML]{333333} \begin{tabular}[c]{@{}c@{}}MAPE\\ var\end{tabular}} \\ \hline
\rowcolor[HTML]{FFFFFF} 
\multicolumn{1}{l|}{\cellcolor[HTML]{FFFFFF}{\color[HTML]{333333} DCRNN}}          & 19.4362                                                                            & 39.03\%                                & 30.3832                                & 545.3596                                                                 & \multicolumn{1}{c|}{\cellcolor[HTML]{FFFFFF}0.5182}                                                                    & 24.6547                                & 55.71\%                                & 37.3885                                & 790.0374                                                                 & \multicolumn{1}{c|}{\cellcolor[HTML]{FFFFFF}1.2729}                                                                    & 32.7574                                & 86.90\%                                & 47.4751                                & 1180.8402                                                                & 3.3537                                                                    \\
\rowcolor[HTML]{FFFFFF} 
\multicolumn{1}{l|}{\cellcolor[HTML]{FFFFFF}{\color[HTML]{333333} STGCN}}          & 13.7359                                                                            & 40.48\%                                & 20.6039                                & 235.8448                                                                 & \multicolumn{1}{c|}{\cellcolor[HTML]{FFFFFF}0.4574}                                                                    & 16.2138                                & 51.05\%                                & 23.4688                                & 287.8995                                                                 & \multicolumn{1}{c|}{\cellcolor[HTML]{FFFFFF}0.6644}                                                                    & 27.7782                                & 87.89\%                                & 35.0219                                & 454.9049                                                                 & 1.8565                                                                    \\
\rowcolor[HTML]{FFFFFF} 
\multicolumn{1}{l|}{\cellcolor[HTML]{FFFFFF}{\color[HTML]{333333} AGCRN}}          & 14.9210                                                                            & 49.35\%                                & 23.9165                                & 349.5797                                                                 & \multicolumn{1}{c|}{\cellcolor[HTML]{FFFFFF}0.7642}                                                                    & 17.2767                                & 60.84\%                                & 26.7793                                & 418.9212                                                                 & \multicolumn{1}{c|}{\cellcolor[HTML]{FFFFFF}1.2171}                                                                    & 19.7528                                & 73.67\%                                & 29.7286                                & 493.9549                                                                 & 1.7813                                                                    \\
\rowcolor[HTML]{FFFFFF} 
\multicolumn{1}{l|}{\cellcolor[HTML]{FFFFFF}{\color[HTML]{333333} MTGNN}}          & {\ul 11.9372}                                                                      & 38.93\%                                & 19.3084                                & 230.4142                                                                 & \multicolumn{1}{c|}{\cellcolor[HTML]{FFFFFF}0.4255}                                                                    & 14.3814                                & 47.84\%                                & {\ul 22.9801}                          & 321.3987                                                                 & \multicolumn{1}{c|}{\cellcolor[HTML]{FFFFFF}{\ul 0.6921}}                                                              & {\ul 17.0799}                          & {\ul 59.89\%}                          & {\ul 26.5632}                          & {\ul 414.0658}                                                           & {\ul 1.1253}                                                              \\
\rowcolor[HTML]{FFFFFF} 
\multicolumn{1}{l|}{\cellcolor[HTML]{FFFFFF}{\color[HTML]{333333} D$^2$STGNN}}        & 12.0410                                                                            & {\ul 37.85\%}                          & \textbf{19.0750}                       & 228.8712                                                                 & \multicolumn{1}{c|}{\cellcolor[HTML]{FFFFFF}0.4370}                                                                    & 14.7446                                & 48.29\%                                & 23.1617                                & 319.0642                                                                 & \multicolumn{1}{c|}{\cellcolor[HTML]{FFFFFF}0.7677}                                                                    & 18.3538                                & 63.46\%                                & 28.3847                                & 468.8282                                                                 & 1.3700                                                                    \\
\rowcolor[HTML]{FFFFFF} 
\multicolumn{1}{l|}{\cellcolor[HTML]{FFFFFF}{\color[HTML]{333333} ST-SSL}}         & 11.8982                                                                            & 38.78\%                                & {\ul 19.2251}                          & 228.0394                                                                 & \multicolumn{1}{c|}{\cellcolor[HTML]{FFFFFF}0.4399}                                                                    & {\ul 14.2621}                          & {\ul 47.54\%}                          & \textbf{22.5824}                       & 316.5560                                                                 & \multicolumn{1}{c|}{\cellcolor[HTML]{FFFFFF}0.6987}                                                                    & 17.1275                                & 62.69\%                                & 26.6144                                & 414.9765                                                                 & 1.2057                                                                    \\ \hline
\rowcolor[HTML]{FFFFFF} 
\multicolumn{1}{l|}{\cellcolor[HTML]{FFFFFF}{\color[HTML]{333333} FAirST+MTGNN}}   & \textbf{11.8854}                                                                   & \textbf{37.03\%}                       & 19.2816                                & \textbf{225.6162}                                                        & \multicolumn{1}{c|}{\cellcolor[HTML]{FFFFFF}\textbf{0.3517}}                                                           & \textbf{14.1500}                       & \textbf{46.20\%}                       & 23.1401                                & \textbf{305.4699}                                                        & \multicolumn{1}{c|}{\cellcolor[HTML]{FFFFFF}\textbf{0.0600}}                                                           & \textbf{16.7561}                       & \textbf{58.98\%}                       & \textbf{26.0058}                       & \textbf{395.7106}                                                        & \textbf{1.0603}                                                           \\
\rowcolor[HTML]{FFFFFF} 
\multicolumn{1}{l|}{\cellcolor[HTML]{FFFFFF}{\color[HTML]{333333} FAirST+D$^2$STGNN}} & 12.4411                                                                            & 39.86\%                                & 19.4670                                & {\ul 226.3241}                                                           & \multicolumn{1}{c|}{\cellcolor[HTML]{FFFFFF}{\ul 0.4174}}                                                              & 14.8216                                & 48.49\%                                & 23.1093                                & {\ul 314.5615}                                                           & \multicolumn{1}{c|}{\cellcolor[HTML]{FFFFFF}0.7352}                                                                    & 17.8283                                & 60.69\%                                & 27.5562                                & 441.7820                                                                 & 1.2329                                                                    \\ \hline
\rowcolor[HTML]{FFFFFF} 
\multicolumn{16}{c}{\cellcolor[HTML]{FFFFFF}ETT} \\ \hline
\rowcolor[HTML]{FFFFFF} 
\multicolumn{1}{l|}{\cellcolor[HTML]{FFFFFF}{\color[HTML]{333333} }}               & \multicolumn{5}{c|}{\cellcolor[HTML]{FFFFFF}{\color[HTML]{333333} horizon-3}} & \multicolumn{5}{c|}{\cellcolor[HTML]{FFFFFF}{\color[HTML]{333333} horizon-6}}  & \multicolumn{5}{c}{\cellcolor[HTML]{FFFFFF}{\color[HTML]{333333} horizon-12}} \\ \cline{2-16} 
\rowcolor[HTML]{FFFFFF} 
\multicolumn{1}{l|}{\cellcolor[HTML]{FFFFFF}{\color[HTML]{333333} }}               & {\color[HTML]{333333} MAE}                                                         & {\color[HTML]{333333} MAPE}            & {\color[HTML]{333333} RMSE}            & {\color[HTML]{333333} \begin{tabular}[c]{@{}c@{}}MAE\\ var\end{tabular}} & \multicolumn{1}{c|}{\cellcolor[HTML]{FFFFFF}{\color[HTML]{333333} \begin{tabular}[c]{@{}c@{}}MAPE\\ var\end{tabular}}} & {\color[HTML]{333333} MAE}             & {\color[HTML]{333333} MAPE}            & {\color[HTML]{333333} RMSE}            & {\color[HTML]{333333} \begin{tabular}[c]{@{}c@{}}MAE\\ var\end{tabular}} & \multicolumn{1}{c|}{\cellcolor[HTML]{FFFFFF}{\color[HTML]{333333} \begin{tabular}[c]{@{}c@{}}MAPE\\ var\end{tabular}}} & {\color[HTML]{333333} MAE}             & {\color[HTML]{333333} MAPE}            & {\color[HTML]{333333} RMSE}            & {\color[HTML]{333333} \begin{tabular}[c]{@{}c@{}}MAE\\ var\end{tabular}} & {\color[HTML]{333333} \begin{tabular}[c]{@{}c@{}}MAPE\\ var\end{tabular}} \\ \hline
\rowcolor[HTML]{FFFFFF} 
\multicolumn{1}{l|}{\cellcolor[HTML]{FFFFFF}{\color[HTML]{333333} DCRNN}}          & 2.0154                                                                             & 21.03\%                                & 3.0606                                 & 6.8057                                                                   & \multicolumn{1}{c|}{\cellcolor[HTML]{FFFFFF}0.1713}                                                                    & 2.5277                                 & 26.65\%                                & 4.8280                                 & 10.5346                                                                  & \multicolumn{1}{c|}{\cellcolor[HTML]{FFFFFF}0.3474}                                                                    & 2.9636                                 & 30.76\%                                & 4.3930                                 & 13.5997                                                                  & 0.5401                                                                    \\
\rowcolor[HTML]{FFFFFF} 
\multicolumn{1}{l|}{\cellcolor[HTML]{FFFFFF}{\color[HTML]{333333} STGCN}}          & 3.7973                                                                             & 47.16\%                                & 5.0410                                 & 10.9922                                                                  & \multicolumn{1}{c|}{\cellcolor[HTML]{FFFFFF}0.5943}                                                                    & 3.4983                                 & 30.87\%                                & 11.0721                                & 3.4983                                                                   & \multicolumn{1}{c|}{\cellcolor[HTML]{FFFFFF}0.1272}                                                                    & 3.1639                                 & 25.57\%                                & 4.5034                                 & 10.2705                                                                  & 0.0789                                                                    \\
\rowcolor[HTML]{FFFFFF} 
\multicolumn{1}{l|}{\cellcolor[HTML]{FFFFFF}{\color[HTML]{333333} AGCRN}}          & 1.4100                                                                             & 9.14\%                                 & 2.0400                                 & 2.2083                                                                   & \multicolumn{1}{c|}{\cellcolor[HTML]{FFFFFF}0.0275}                                                                    & 1.6100                                 & 10.32\%                                & 2.3900                                 & 3.1247                                                                   & \multicolumn{1}{c|}{\cellcolor[HTML]{FFFFFF}0.0249}                                                                    & 1.7500                                 & 13.15\%                                & 2.6177                                 & 3.9166                                                                   & 0.0250                                                                    \\
\rowcolor[HTML]{FFFFFF} 
\multicolumn{1}{l|}{\cellcolor[HTML]{FFFFFF}{\color[HTML]{333333} MTGNN}}          & 1.5023                                                                             & 12.81\%                                & 2.1853                                 & 2.5811                                                                   & \multicolumn{1}{c|}{\cellcolor[HTML]{FFFFFF}0.0267}                                                                    & 1.8136                                 & 14.45\%                                & 2.7287                                 & 4.2237                                                                   & \multicolumn{1}{c|}{\cellcolor[HTML]{FFFFFF}0.0304}                                                                    & 1.9305                                 & 15.44\%                                & 2.9455                                 & 4.8330                                                                   & 0.0392                                                                    \\
\rowcolor[HTML]{FFFFFF} 
\multicolumn{1}{l|}{\cellcolor[HTML]{FFFFFF}{\color[HTML]{333333} D$^2$STGNN}}        & \textbf{1.2199}                                                                    & \textbf{10.35\%}                       & \textbf{1.8181}                        & {\ul 1.9608}                                                             & \multicolumn{1}{c|}{\cellcolor[HTML]{FFFFFF}0.0181}                                                                    & \textbf{1.5175}                        & \textbf{12.27\%}                       & \textbf{2.1841}                        & {\ul 2.8447}                                                             & \multicolumn{1}{c|}{\cellcolor[HTML]{FFFFFF}0.0222}                                                                    & {\ul 1.7474}                           & 13.38\%                                & {\ul 2.6303}                           & {\ul 3.9519}                                                             & 0.0231                                                                    \\
\rowcolor[HTML]{FFFFFF} 
\multicolumn{1}{l|}{\cellcolor[HTML]{FFFFFF}{\color[HTML]{333333} ST-SSL}}         & 1.4541                                                                             & 12.83\%                                & 2.0891                                 & 2.3073                                                                   & \multicolumn{1}{c|}{\cellcolor[HTML]{FFFFFF}0.0265}                                                                    & 1.9052                                 & 14.19\%                                & 2.8121                                 & 4.3810                                                                   & \multicolumn{1}{c|}{\cellcolor[HTML]{FFFFFF}0.0189}                                                                    & 1.9843                                 & 14.97\%                                & 2.8741                                 & 4.4265                                                                   & 0.0247                                                                    \\ \hline
\rowcolor[HTML]{FFFFFF} 
\multicolumn{1}{l|}{\cellcolor[HTML]{FFFFFF}{\color[HTML]{333333} FAirST+MTGNN}}   & 1.3625                                                                             & 10.95\%                                & 2.0019                                 & 2.2038                                                                   & \multicolumn{1}{c|}{\cellcolor[HTML]{FFFFFF}\textbf{0.0146}}                                                           & 1.6359                                 & 12.49\%                                & 2.4444                                 & 3.3768                                                                   & \multicolumn{1}{c|}{\cellcolor[HTML]{FFFFFF}{\ul 0.0187}}                                                              & 1.7832                                 & {\ul 12.89\%}                          & 2.7225                                 & 4.3247                                                                   & {\ul 0.0223}                                                              \\
\rowcolor[HTML]{FFFFFF} 
\multicolumn{1}{l|}{\cellcolor[HTML]{FFFFFF}{\color[HTML]{333333} FAirST+D$^2$STGNN}} & {\ul 1.3081}                                                                       & {\ul 10.92\%}                          & {\ul 1.8956}                           & \textbf{1.9292}                                                          & \multicolumn{1}{c|}{\cellcolor[HTML]{FFFFFF}{\ul 0.0155}}                                                              & {\ul 1.5397}                           & {\ul 12.33\%}                          & {\ul 2.2631}                           & \textbf{2.8174}                                                          & \multicolumn{1}{c|}{\cellcolor[HTML]{FFFFFF}\textbf{0.0185}}                                                           & \textbf{1.6725}                        & \textbf{12.77\%}                       & \textbf{2.5542}                        & \textbf{3.8085}                                                          & \textbf{0.0201}                                                           \\ \hline
\end{tabular}
}
\end{center}
\end{table*}

\textbf{Performance on fairness metrics}. Our FairSTG significantly surpasses all other baselines on  fairness metrics across all tasks. Specifically, for the farthest  step prediction, the MAE-var improvements at horizon-12  over the best baseline are 5.59\%, 8.15\%, 4.43\% and 2.76\% on METR-LA, PEMS-BAY, KnowAir and ETT, respectively, and the  improvements on MAPE-var  are 12.36\%, 12.01\%, 5.77\% and 12.98\% on METR-LA, PEMS-BAY, KnowAir and ETT. We attribute such superiority to two aspects, i.e., 1) Baseline models primarily focus on improving overall performance, inducing the  overfitting of samples with  lower difficulty and neglecting the challenging subgroup.  2) Our FairSTG is equipped with advantageous compensation mechanism and introduces fairness constraints on both representation space and optimization objective, thereby enhancing the model's fairness performance.

\textbf{Performance on forecasting metrics.} We observe that our FairSTG framework achieves comparable forecasting accuracy in all datasets, noting that there usually exists a trade-off between fairness and accuracy ~\cite{bose2019compositional, dai2021say, liu2023generalized}. Particularly, our FairSTG  achieves optimal forecasting performance at horizon-12 on all datasets, where the superiority can be interpreted as two aspects. First, 
as the forecasting horizon increases, the regularity between predictions and inputs weakens and the learning difficulty increases. Thus, learning without fairness can be trapped into a lazy learning mode, directly neglecting the samples with higher learning difficulty.
Secondly, in addition to  fairness-aware learning objective, our FairSTG is capable 
of suppressing the prediction accuracy  diversity via the joint optimization of sample-level representation. i.e., collaborative feature enhancement and adaptive mix-up for fair representation. Therefore, our  FairSTG reasonably achieves superior performance on both evaluations of accuracy and fairness.
%as the forecasting horizon increases, the regularity between predictions and inputs becomes weaken with learning difficulty increasing. In pursuit of superior overall performance, spatiotemporal learning models are more likely to capture patterns with lower learning difficulty, while inadvertently neglects patterns with higher learning difficulty, thus generating inferior and unfair predictions. 

\begin{figure*}[htbp]
    \centering
    \includegraphics[width=\textwidth]{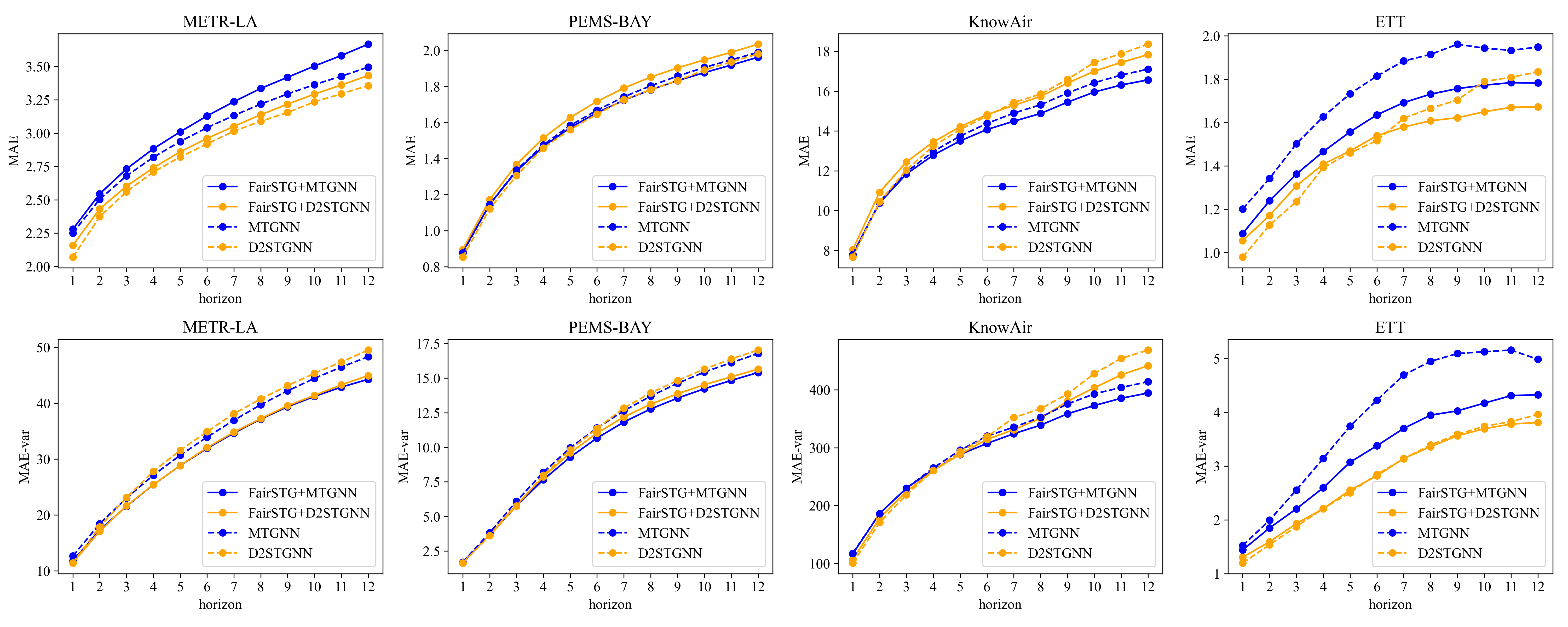}
    \caption{Forecasting and fairness performance comparison at each horizon. The top  and bottom lines respectively indicate the prediction accuracy and fairness performances.}
    \label{fig:trend}
\end{figure*}

\textbf{Generality and plug-ability of FairSTG.} We further compare the empirical prediction performances of our FairSTG  with two backbone models. As shown in Fig ~\ref{fig:trend}, the fairness metrics are consistently superior to backbones along with the increases of forecasting horizons. Besides, we observe that our FairSTGs carrying with different backbones can  achieve similar and comparable performance with corresponding backbones, illustrating the stability of the whole architecture of of FairSTG. It is worthnoting that our FairSTG achieves optimal performance in both KnowAir and ETT datasets. The reason behind this phenomenon lies in that these two datasets have relatively small scales, where the collaborative feature enhancement of FairSTG can exactly alleviate data sparsity and improve the overall performance. Such satisfactory joint enhancement further validates the intuition that mitigating forecasting heterogeneity contributes to the promotion of overall performance. In summary, our integrated FairSTG can  enable the fairness  for  different backbone models and maintain comparable forecasting performance, which verifies the generality of our FairSTG framework. 

\subsection{The improvement in challenging samples}

%Our FairSTG framework focuses on mitigating the unfairness issue in spatiotemporal forecasting, and collaboratively enhances the model's forecasting performance on the challenging subgroup at both the representation level and objective level. Following the definition of easy samples and challenging samples in Definition 3, we conduct a comparison experiment on easy set and challenging set at horiozon 12.
One of the goals of our FairSTG is to alleviate the  unfairness on challenging samples which are difficult to learn, thus it is necessary to verify whether FairSTG improves the forecasting performance on such subgroup. For an explicit comparison, we select samples with performance at the top 30\% as easy ones, and samples with performance at the bottom 30\% as challenging ones. We then compare the integrated FairSTGs with corresponding vanilla backbones.  As shown in Table ~\ref{tab:challenging}, our FairSTG outperforms all baseline models on challenging set at both accuracy metric MAE and  fairness metric MAE-var, while maintaining comparable performance on easy set  with satisfactory MAE and MAE-var. 
Additionally, for each group learned by FairSTG, performances on the challenging set overwhelmingly  surpass the corresponding backbone models on accuracy metric, i.e., MAE, which refers to that our FairSTG indeed improves model's expressive capacity on challenging samples which are prone to suffer unfair treatment, and thus mitigating the forecasting heterogeneity and unfairness issue in spatiotemporal learning. 
Moreover, on datasets of KnowAir and ETT, our FairSTG  simultaneously achieves optimal results on  forecasting accuracy and fairness metrics, on both easy set and challenging set. As analyzed earlier, since KnowAir and ETT  are with a relatively small scale, the collaborative feature enhancement in our FairSTG can alleviate the data sparsity and thus improving the overall forecasting and fairness performance. 
In brief, through the analysis on inter-group and intra-group prediction performance and fairness learning quality, on  respective easy set and challenging set, we can further verify that FairSTG is equipped with the capacity of alleviating unfairness on challenging samples. Therefore, our FairSTG can exactly compensate  the samples with lower regularity and under-representation, potentially suppressing the discrimination in urban intelligent system.

\begin{table}[]
\caption{\linebreak The comparison on easy set and challenging set.}
\label{tab:challenging}
\resizebox{0.45\textwidth}{!}{
\begin{tabular}{lcccc}
\hline
\multicolumn{5}{c}{METR-LA}                                                                                                                                                                                                                                               \\ \hline
                & \begin{tabular}[c]{@{}c@{}}easy set\\ MAE\end{tabular} & \begin{tabular}[c]{@{}c@{}}easy set\\ MAE-var\end{tabular} & \begin{tabular}[c]{@{}c@{}}challenging set\\ MAE\end{tabular} & \begin{tabular}[c]{@{}c@{}}challenging set\\ MAE-var\end{tabular} \\ \hline
MTGNN           & 0.1687                                                 & 0.0355                                                     & 10.0551                                                        & 105.2392                                                          \\
D$^2$STGNN         & \textbf{0.1026}                                        & \textbf{0.0201}                                            & {\ul 9.5583}                                                  & 112.0150                                                          \\
FairSTG+MTGNN   & 0.1834                                                 & 0.0419                                                     & 9.9270                                                        & \textbf{95.2383}                                                  \\
FairSTG+D$^2$STGNN & {\ul 0.1102}                                           & {\ul 0.0231}                                               & \textbf{9.4708}                                               & {\ul 96.2490}                                                  \\ \hline
\multicolumn{5}{c}{PEMS-BAY}                                                                                                                                                                                                                                              \\ \hline
                & \begin{tabular}[c]{@{}c@{}}easy set\\ MAE\end{tabular} & \begin{tabular}[c]{@{}c@{}}easy set\\ MAE-var\end{tabular} & \begin{tabular}[c]{@{}c@{}}challenging set\\ MAE\end{tabular} & \begin{tabular}[c]{@{}c@{}}challenging set\\ MAE-var\end{tabular} \\ \hline
MTGNN           & {\ul 0.2111}                                           & {\ul 0.0160}                                               & 5.2211                                                        & 40.5645                                                           \\
D$^2$STGNN         & \textbf{0.2037}                                        & \textbf{0.0153}                                            & 5.3478                                                        & 41.3632                                                           \\
FairSTG+MTGNN   & 0.2176                                                 & 0.0169                                                     & \textbf{5.1770}                                               & {\ul 37.1559}                                                     \\
FairSTG+D$^2$STGNN & 0.2273                                                 & 0.0185                                                     & {\ul 5.2904}                                                  & \textbf{36.5698}                                                  \\ \hline
\multicolumn{5}{c}{KnowAir}                                                                                                                                                                                                                                               \\ \hline
                & \begin{tabular}[c]{@{}c@{}}easy set\\ MAE\end{tabular} & \begin{tabular}[c]{@{}c@{}}easy set\\ MAE-var\end{tabular} & \begin{tabular}[c]{@{}c@{}}challenging set\\ MAE\end{tabular} & \begin{tabular}[c]{@{}c@{}}challenging set\\ MAE-var\end{tabular} \\ \hline
MTGNN           & {\ul 3.1115}                                           & {\ul 3.3080}                                                & {\ul 38.2072}                                                 & 679.0028                                                          \\
D$^2$STGNN         & 3.2667                                                 & 3.6180                                                      & 41.5705                                                       & 724.8887                                                          \\
FairSTG+MTGNN   & \textbf{3.0611}                                        & \textbf{3.2038}                                            & \textbf{36.7819}                                              & \textbf{669.3046}                                                 \\
FairSTG+D$^2$STGNN & 3.1327                                                 & 3.3571                                                     & 40.4428                                                       & {\ul 677.5437}                                                    \\ \hline
\multicolumn{5}{c}{ETT}                                                                                                                                                                                                                                                   \\ \hline
                & \begin{tabular}[c]{@{}c@{}}easy set\\ MAE\end{tabular} & \begin{tabular}[c]{@{}c@{}}easy set\\ MAE-var\end{tabular} & \begin{tabular}[c]{@{}c@{}}challenging set\\ MAE\end{tabular} & \begin{tabular}[c]{@{}c@{}}challenging set\\ MAE-var\end{tabular} \\ \hline
MTGNN           & 0.2987                                                 & 0.0369                                                     & 4.4653                                                        & 6.5106                                                            \\
D$^2$STGNN         & 0.3179                                                 & 0.0412                                                     & 4.0504                                                        & 5.0291                                                            \\
FairSTG+MTGNN   & \textbf{0.2648}                                        & \textbf{0.0268}                                            & {\ul 4.0310}                                                  & \textbf{5.4197}                                                   \\
FairSTG+D$^2$STGNN & {\ul 0.2657}                                           & {\ul 0.0292}                                               & \textbf{3.8328}                                               & {\ul 5.4260}                                                      \\ \hline
\end{tabular}
}
\end{table}

\subsection{The analysis of fairness recognizer}
\begin{figure}[htbp]
    \centering
    \includegraphics[width=0.75\linewidth]{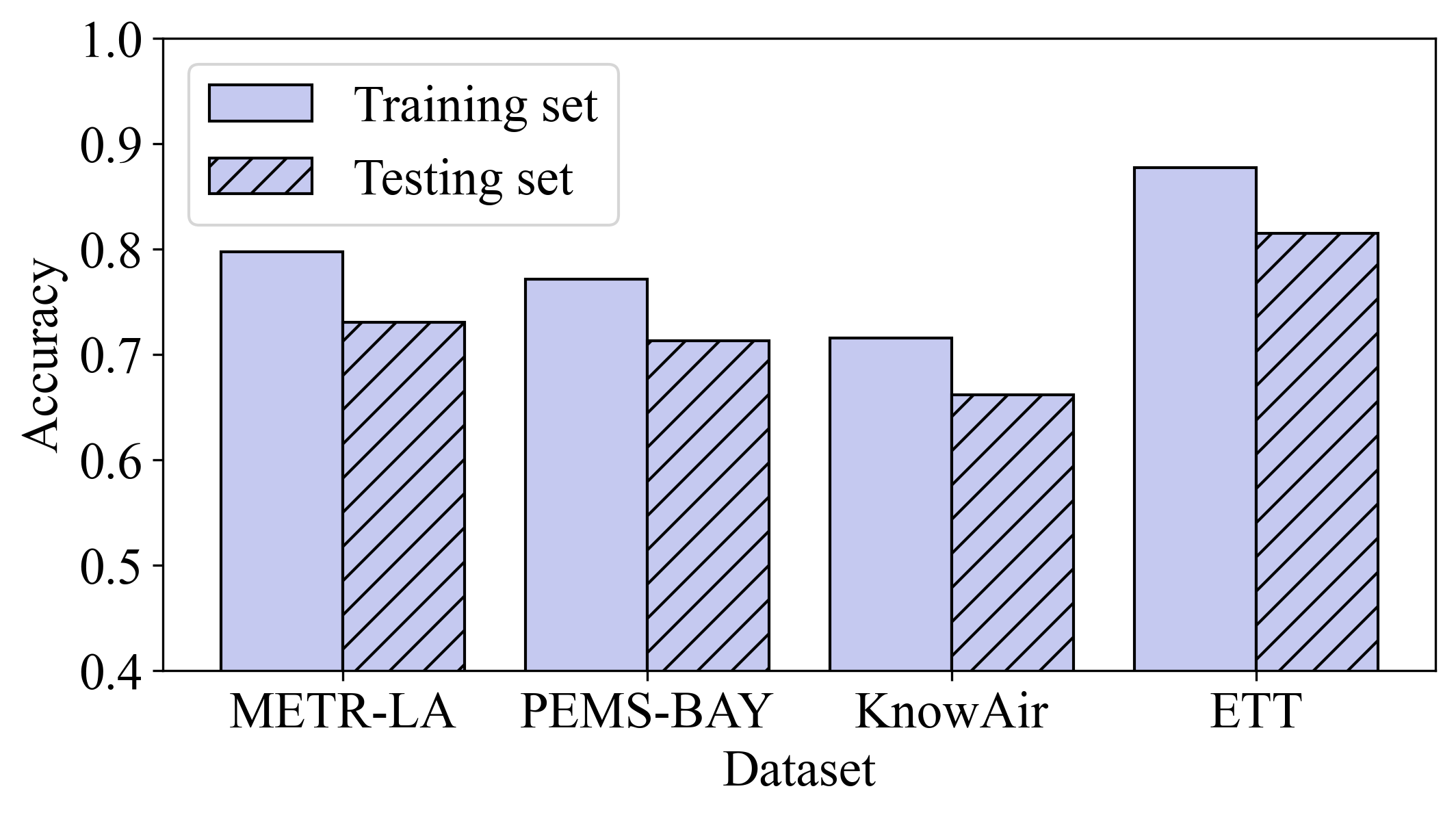}
    % \captionsetup{width=0.9\linewidth}
    \caption{The accuracy of fairness recognizer in self-supervised classification.}
    \label{fig:acc}
\end{figure}

\textbf{The effectiveness of the fairness recognizer.} We first verify the correctness of our intuition, i.e., the learning difficulty of a spatiotemporal sample should be associated with its sequential features and external factors. Even though spatiotemporal datasets lack explicit supervision of annotations on challenging samples,  we design a fairness recognizer,  to learn the difficulty of different spatiotemporal samples during training and make inference during testing. In Fig.~\ref{fig:acc}, we report the accuracy of the self-supervised binary classification task in FairSTG with the backbone D$^2$STGNN. 
The accuracy of fairness recognizer  ranges from  72.20\% to 87.40\% on the training set, while accounts for the ranges from  66.19\% to  81.49\% on the test set. The  accurate and high-quality predictions  on all four datasets   indicate that challenging samples can be well-identified in the absence of explicitly labeled sensitive attributes and can exactly satisfy further fair predictions.  

\textbf{The architecture of fairness recognizer.} We investigate how the neural network architecture influences the performance of fairness recognizer.
In our main experiments, we utilize a three-layer  GCN  as the architecture to capture the spatiotemporal correlations. We then take a linear three-layer MLP as an alternative for the impact analysis, where the initialized and modified one are designated as FairSTG-GCN(3) and FairSTG-Linear(3). Our implementation is based on D$^2$STGNN at horizon = $12$ and the results are reported in Table ~\ref{tab: fair_architecture}. In terms of the accuracy of self-supervised classification, FairSTG-GCN(3) surpasses FairSTG-Linear(3) in almost all scenarios. This is probably because the GCN structure excels in capturing correlations in graph structures, thus  perceiving the learning difficulty of different  samples more accurately. Moreover, for forecasting performance and fairness metrics, FairSTG-GCN(3) slightly outperforms FairSTG-Linear(3), which further delivers that the GCN-based fairness recognizer should be more suitable for fair spatiotemporal forecasting as the inherent regularity of observations are associated with  underlying spatial dependencies.

\begin{table}[htbp]
% \captionsetup{width=0.9\linewidth}
\caption{The results of different architectures for the fairness recognizer.}
\label{tab: fair_architecture}
\begin{tabular}{lcccc}
\hline
\multicolumn{5}{c}{METR-LA}                                                                                                                                                                                                                                     \\ \hline
\multicolumn{1}{l|}{}                & \multicolumn{1}{l}{MAE} & \multicolumn{1}{l}{MAE-var} & \multicolumn{1}{l}{\begin{tabular}[c]{@{}l@{}}Training\\ accuracy\end{tabular}} & \multicolumn{1}{l}{\begin{tabular}[c]{@{}l@{}}Testing\\ accuracy\end{tabular}} \\ \hline
\multicolumn{1}{l|}{FairSTG-GCN(3)}    & \textbf{3.4316}                  & \textbf{44.9188}                     & \textbf{0.7972}                                                                          & 0.7309                                                                         \\
\multicolumn{1}{l|}{FairSTG-Linear(3)} & 3.4509                  & 46.5286                     & 0.7860                                                                          & \textbf{0.7471}                                                                         \\ \hline
\multicolumn{5}{c}{PEMS-BAY}                                                                                                                                                                                                                                    \\ \hline
\multicolumn{1}{l|}{}                & \multicolumn{1}{l}{MAE} & \multicolumn{1}{l}{MAE-var} & \multicolumn{1}{l}{\begin{tabular}[c]{@{}l@{}}Training\\ accuracy\end{tabular}} & \multicolumn{1}{l}{\begin{tabular}[c]{@{}l@{}}Testing\\ accuracy\end{tabular}} \\ \hline
\multicolumn{1}{l|}{FairSTG-GCN(3)}    & \textbf{2.0349}                  & \textbf{15.6454}                     & \textbf{0.7714}                                                                          & 0.7131                                                                         \\
\multicolumn{1}{l|}{FairSTG-Linear(3)} & 2.0352                  & 15.9652                     & 0.7171                                                                          & \textbf{0.7296}                                                                         \\ \hline
\multicolumn{5}{c}{KnowAir}                                                                                                                                                                                                                                     \\ \hline
\multicolumn{1}{l|}{}                & \multicolumn{1}{l}{MAE} & \multicolumn{1}{l}{MAE-var} & \multicolumn{1}{l}{\begin{tabular}[c]{@{}l@{}}Training\\ accuracy\end{tabular}} & \multicolumn{1}{l}{\begin{tabular}[c]{@{}l@{}}Testing\\ accuracy\end{tabular}} \\ \hline
\multicolumn{1}{l|}{FairSTG-GCN(3)}    & \textbf{17.8283}                 & \textbf{441.7820}                    & \textbf{0.7159}                                                                          & \textbf{0.6619}                                                                         \\
\multicolumn{1}{l|}{FairSTG-Linear(3)} & 18.356                  & 460.5098                    & 0.7054                                                                          & 0.625                                                                          \\ \hline
\multicolumn{5}{c}{ETT}                                                                                                                                                                                                                                         \\ \hline
\multicolumn{1}{l|}{}                & \multicolumn{1}{l}{MAE} & \multicolumn{1}{l}{MAE-var} & \multicolumn{1}{l}{\begin{tabular}[c]{@{}l@{}}Training\\ accuracy\end{tabular}} & \multicolumn{1}{l}{\begin{tabular}[c]{@{}l@{}}Testing\\ accuracy\end{tabular}} \\ \hline
\multicolumn{1}{l|}{FairSTG-GCN(3)}    & \textbf{1.6725}                  & \textbf{3.8085}                      & 0.8774                                                                          & \textbf{0.8149}                                                                         \\
\multicolumn{1}{l|}{FairSTG-Linear(3)} & 1.7056                  & 4.0518                      & \textbf{0.8779}                                                                          & 0.7852                                                                         \\ \hline
\end{tabular}
\end{table}

\subsection{Case study}
In this subsection, we investigate how compensatory samples generate and collaboratively enhance the representation of challenging-to-learn  samples through a toy case. 

\begin{figure}[htbp]
    \centering
    \includegraphics[width=\linewidth]{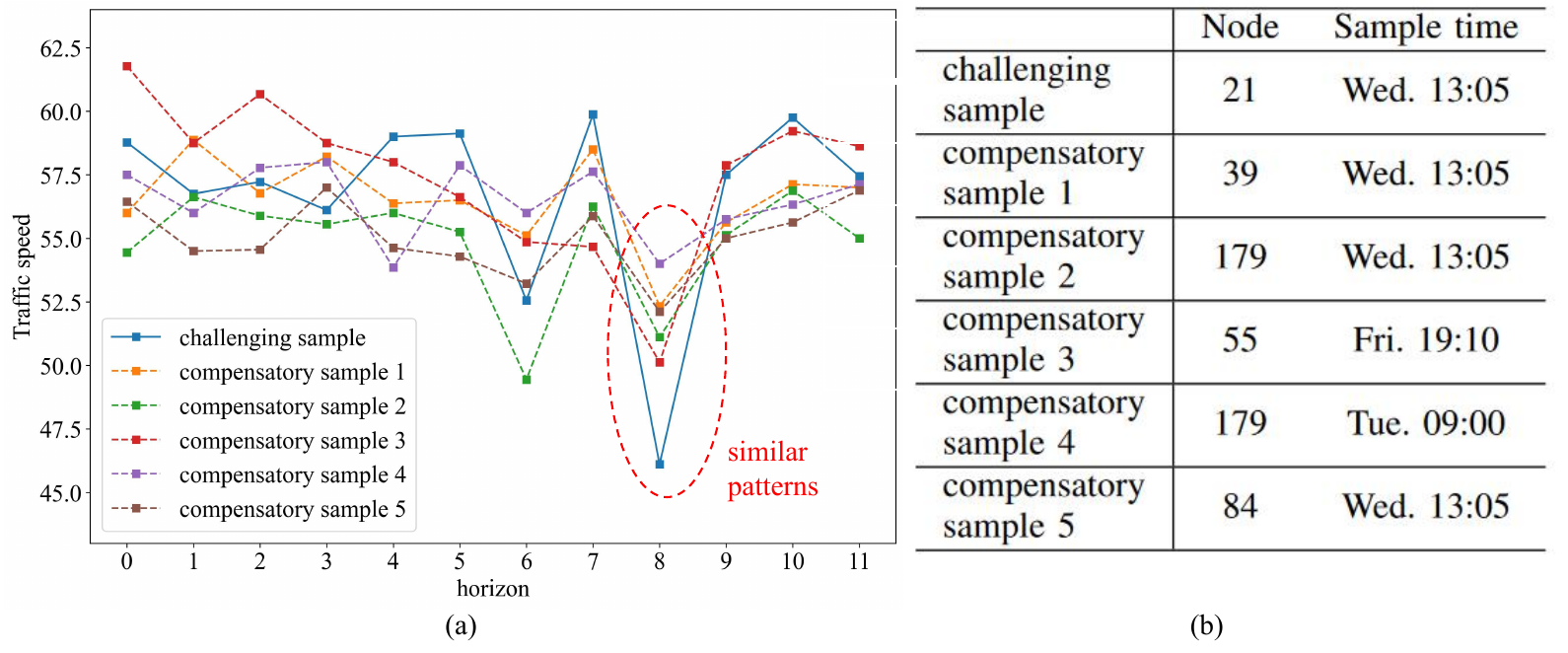}
    \caption{Case study in compensatory samples. (a) The raw time series of a challenging sample and its compensatory samples. (b) The sampling time stamps and nodes for these samples. }
    \label{fig:case_study1}
\end{figure}

\textbf{The generation of compensatory samples.} Given a challenging sample, the compensatory samples are derived by computing the  similarity between challenging ones and well-learned ones, where the compensatory set is extracted from different spatiotemporal graphs. We first plot the original time series of a challenging sample and the generated compensatory samples derived by our FairSTG. As shown in Fig.~\ref{fig:case_study1}(a), all compensatory samples exhibit similar patterns with the challenging one, indicating that our FairSTG   can capture effective samples with in a  high-quality manner. 
Besides, we further investigate the sampling time stamps and nodes of these samples. As shown in Fig.~\ref{fig:case_study1}(b),
for compensatory samples  1, 2, and 5, where they come from the same spatiotemporal graph to the challenging sample, while compensatory samples 3 and 4 are sampled at other time stamps. It delivers that our compensatory sample generation can span the search space of advantageous representations from a single ST Graph to the samples within whole  batch and significantly extends the scope of advantageous representation transfer. Consequently, the model can delve deeper into exploring and leveraging the cross-step spatiotemporal correlations within observations, alleviating the unfairness across temporal perspective. Actually, generating more informative representations from compensatory set for challenging samples can compensate for the insufficiency of model's  expressive power, and further gain both fairness quality and prediction accuracy within  spatiotemporal learning.

\begin{figure}[htbp]
    \centering
    \includegraphics[width=\linewidth]{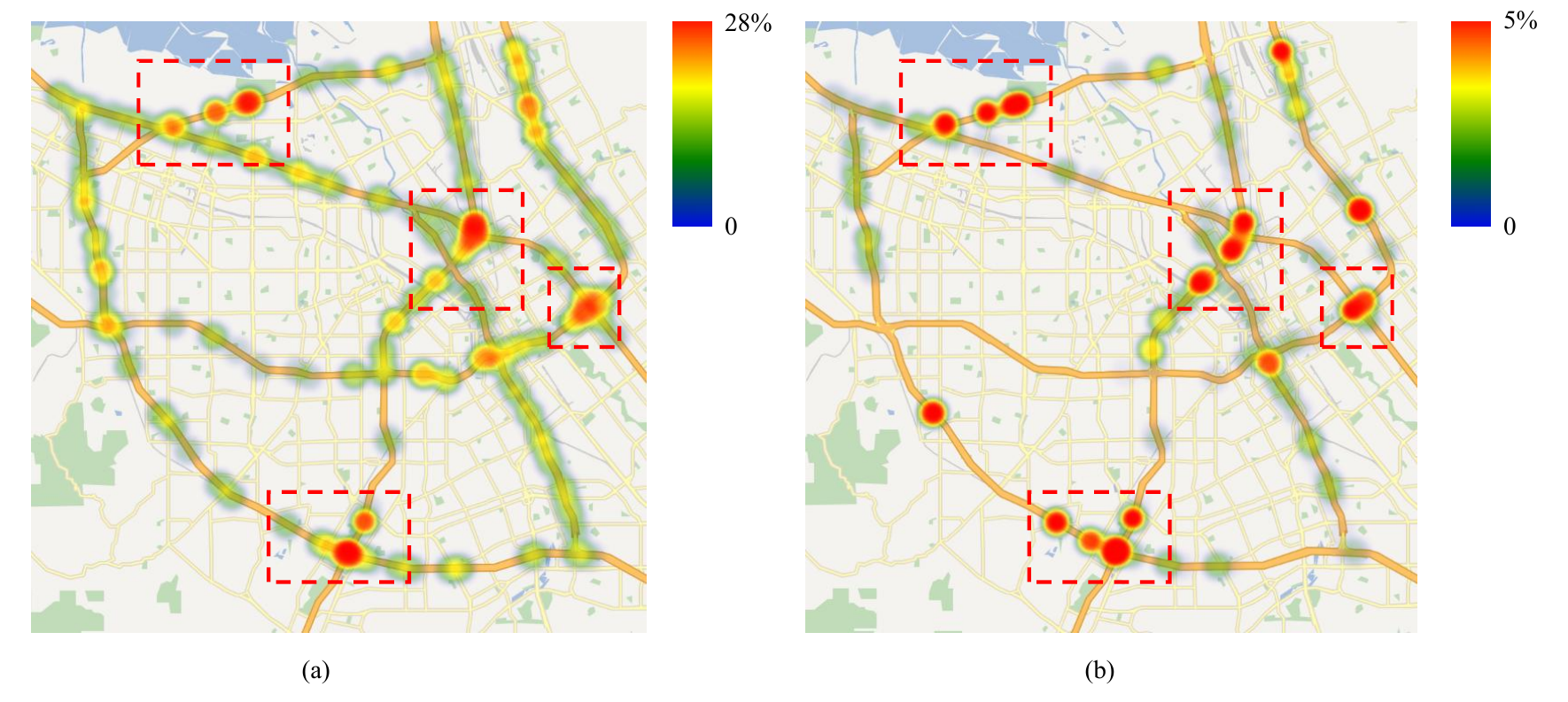}
    \caption{Case study on challenging samples. (a) The forecasting performance (MAPE) of each sensor in PEMS-BAY dataset. (b) The improvement value of each sensor, and only sensors with improvement are plotted.}
    \label{fig:case_study2}
\end{figure}

\textbf{Visualization of performance improvement on challenging samples.} We visualize the spatial distribution of prediction results based on our FairSTG,  to evaluate its capability in  improving performance of the challenging subgroup. We first plot the error (here we exploit MAPE) of each sensor based on  MTGNN in PEMS-BAY, where the red boxes highlight regions with greater errors. In Fig.~\ref{fig:case_study2}(a), it can be observed that sensors located at transportation hubs often suffer more serious prediction errors, possibly due to their  complex traffic patterns and higher learning difficulty, inducing susceptibility to unfair treatment. Additionally, we further visualize the improvement of an error metric in Fig.~\ref{fig:case_study2}(b). To be specific, the  improvement of  MAPE on $i$-th sensor can be illustrated as $\Delta e_i = e_i-e_i'$, where $e_i$ and $e_i'$ respectively indicate the MAPE of the backbone and our FairSTG. Noting that $\Delta e_i > 0$ means FairSTG outperforms the backbone model in making more accurate predictions. As shown,  sensors accounting for significant improvement are also concentrated at transportation hubs, consistent with the spatial distribution of sensors with larger errors. This case delivers us that our FairSTG can distinguish the challenging samples from the spatial perspective by constraining the samples into node-levels, while simultaneously enhance the performance of the backbone model on challenging subgroup. Therefore, in a real-world urban applications, such as traffic status or road risk prediction systems, with our FairSTG, the prediction system can emphasize more on underrepresented urban regions, which reflects more consistency with facts, and allocate the resources in a positively fairness-aware manner.

\subsection{Ablation experiments}
We conduct an ablation study to validate the effectiveness of key components that contribute to the improved outcomes of our proposed framework. We name FairSTG without different components as follows. 1) FairSTG-w/o-FE (FairSTG without feature enhancement). We remove our fairness recognizer and collaborative feature enhancement from FairSTG, and only remain fairness constraints in the optimization objective. 2) FairSTG-w/o-FO (FairSTG without fairness objective). We remove the fairness constraints from the optimization objective in our FairSTG and only  take  the original MAE loss and self-supervised BCE loss as the overall objective.

We choose MTGNN as the backbone and report the accuracy performance and fairness quality in TABLE~\ref{tab:ablation}. We can observe that, 1) The fairness metrics of FairSTG-w/o-FE and FairSTG-w/o-FO both exhibit a decrease, indicating that collaborative feature enhancement at the representation level and fairness constraints at objective level  can work collaboratively to  mitigate the unfairness issue in  backbone models. 2) For a relatively large-scale dataset, such as METR-LA and PEMS-BAY, FairSTG-w/o-FO exhibits a more prominent decrease in fairness quality. This suggests that  such large-scale datasets with strong spatiotemporal correlations inherently reveal heterogeneity and introducing fairness constraints in the optimization objective have further proved to be more effective. 3) In contrast, for relatively small-scale datasets such as KnowAir and ETT, FairSTG-w/o-FE exhibits a noticeable decrease in fairness metrics.  This suggests that 
generating compensatory representations at the representation level plays a pivotal role  in such sparse datasets, where it can be  interpreted as that  collaborative feature enhancement can leverage information beyond a single ST Graph, effectively enhancing the expressive power of challenging samples. Therefore, the two components  can  exactly  contribute to  FairSTG, while different datasets can be more sensitive to  specific strategies, and this observation  also inspires us to  further investigate data-adaptive learning design in future work. 

\begin{table}[]
\caption{Ablation experiments.}
\label{tab:ablation}
\resizebox{0.45\textwidth}{!}{
\begin{tabular}{lcccc}
\hline
\multicolumn{5}{c}{METR-LA}                                                                                                                          \\ \hline
\multicolumn{1}{l|}{}              & MAE                     & MAE-var                     & MAPE                     & MAPE-var                     \\ \hline
\multicolumn{1}{l|}{FairSTG}       & 3.6221                  & \textbf{45.5186}                     & 10.17\%                  & 0.1525                       \\
\multicolumn{1}{l|}{FairSTG-w/o-FE} & 3.6103                  & 45.8562                     & \textbf{10.15}\%                  & \textbf{0.1519}                       \\
\multicolumn{1}{l|}{FairSTG-w/o-FO} & \textbf{3.5379}                  & 48.9587                     & 10.17\%                  & 0.1666                       \\ \hline
\multicolumn{5}{c}{PEMS-BAY}                                                                                                                         \\ \hline
\multicolumn{1}{l|}{}              & \multicolumn{1}{l}{MAE} & \multicolumn{1}{l}{MAE-var} & \multicolumn{1}{l}{MAPE} & \multicolumn{1}{l}{MAPE-var} \\ \hline
\multicolumn{1}{l|}{FairSTG}       & \textbf{1.9628}                  & \textbf{15.4070}                     & \textbf{4.58}\%                   & \textbf{0.0448}                       \\
\multicolumn{1}{l|}{FairSTG-w/o-FE} & 1.9925                  & 15.9100                       & 4.64\%                   & 0.0457                       \\
\multicolumn{1}{l|}{FairSTG-w/o-FO} & 2.0100                  & 16.9199                     & 4.70\%                   & 0.0476                       \\ \hline
\multicolumn{5}{c}{KnowAir}                                                                                                                          \\ \hline
\multicolumn{1}{l|}{}              & \multicolumn{1}{l}{MAE} & \multicolumn{1}{l}{MAE-var} & \multicolumn{1}{l}{MAPE} & \multicolumn{1}{l}{MAPE-var} \\ \hline
\multicolumn{1}{l|}{FairSTG}       & 16.7561                 & \textbf{395.7106}                    & \textbf{58.98\%}                  & \textbf{1.0603}                       \\
\multicolumn{1}{l|}{FairSTG-w/o-FE} & 16.9433                 & 421.5523                    & 59.94\%                  & 1.1146                       \\
\multicolumn{1}{l|}{FairSTG-w/o-FO} & \textbf{16.7029}                 & 402.2793                    & 60.09\%                  & 1.1032                       \\ \hline
\multicolumn{5}{c}{ETT}                                                                                                                              \\ \hline
\multicolumn{1}{l|}{}              & \multicolumn{1}{l}{MAE} & \multicolumn{1}{l}{MAE-var} & \multicolumn{1}{l}{MAPE} & \multicolumn{1}{l}{MAPE-var} \\ \hline
\multicolumn{1}{l|}{FairSTG}       & \textbf{1.7832}                  & \textbf{4.3247}                      & \textbf{12.89\%}                  & \textbf{0.0201}                       \\
\multicolumn{1}{l|}{FairSTG-w/o-FE} & 1.8216                  & 4.5512                      & 13.76\%                  & 0.0288                       \\
\multicolumn{1}{l|}{FairSTG-w/o-FO} & 1.7860                  & 4.3745                      & 13.55\%                   & 0.0267                       \\ \hline
\end{tabular}
}
\end{table}

\subsection{Hyper-parameters analysis}
We conduct the parameter study on two core hyper-parameters in our proposed FairSTG, $\mu_f$ within a range of $\{0.01, 0.1, 0.5, 1.0, 1.5\}$, which controls the proportion of fairness  constraint in the learning objective, and $k_c$ within a range of $\{5,10,20\}$ that defines the number of compensatory samples in collaborative representation enhancement.

\begin{figure}
    \centering

    % 第一个子图
    \begin{subfigure}{0.48\textwidth}
        \centering
        \includegraphics[width=\linewidth]{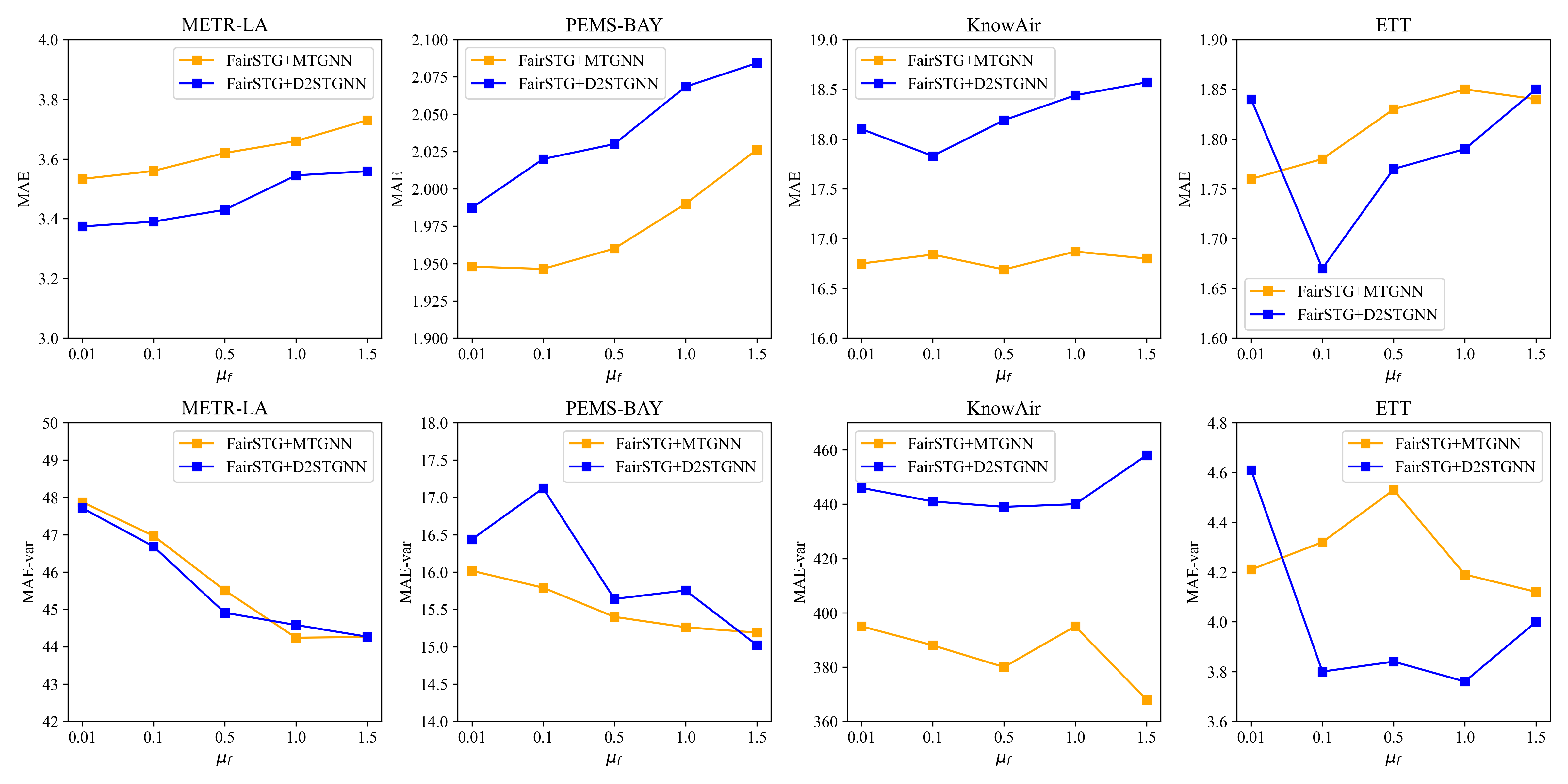}
         \caption{ }
    \end{subfigure}
    % \hspace{0.1\textwidth}  % 适当的水平间距，可以根据需要调整
    % 第二个子图
    \begin{subfigure}{0.48\textwidth}
        \centering
        \includegraphics[width=\linewidth]{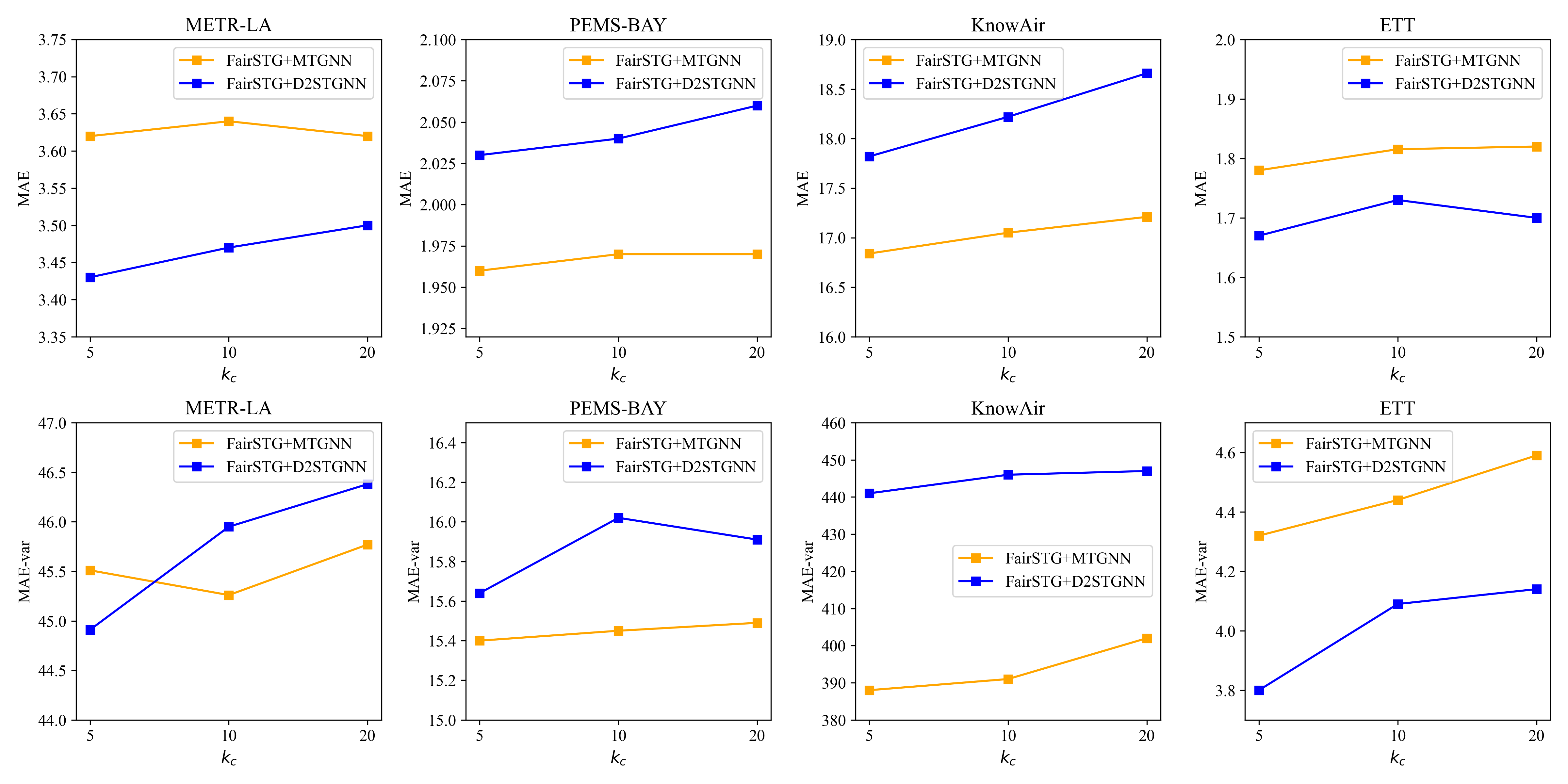}
         \caption{ }
    \end{subfigure}

    \caption{Hyper-parameter study. (a) Performance variation on different $\mu_l$. (b) Performance variation on different $k_c$.}
    \label{fig:hyper}
\end{figure}

We adjust the specific parameter and  fix  the others  in each experiment. Fig. ~\ref{fig:hyper} reports the results of our hyper-parameter study. As shown in Fig. ~\ref{fig:hyper}(a), the forecasting accuracy decreases while  fairness quality  increases  with the increases of  emphasis of fairness $\mu_f$. This implies that the model tends to generate fair predictions for different samples, but at the cost of  decline in overall performance. And it is further verified that there exists an inevitable trade-off between fairness and accuracy in fairness-aware learning. Besides, Fig. ~\ref{fig:hyper}(b) demonstrates that a small number of compensatory samples will result  in better forecasting performance and fairness degree. It is rational  because increasing the number of compensatory samples will unavoidably introduce noises to the mix-up representations and leads to a performance degradation. Overall, we are required to adjust the parameters to achieve a tradeoff between the fairness and accuracy.

\section{Relationship among mobile computing, fairness issue and potential solutions}
In this section, we review the main goal of mobile computing techniques and subsequently raise the fairness issue between urban development and mobile computing techniques. Mobile computing techniques aims to facilitate the data collection, transmission, and exploitation across mobile devices, from the bottom hardware to up applications. Actually, location-based services, such as POI recommendation, fine-grained air quality prediction and traffic resource allocations, are typically applications in mobile computing community where they are highly dependent on spatiotemporal data and corresponding learning algorithms. With the increasing pace of urbanization, the infrastructures and available devices between urban and suburban areas are imbalanced and such imbalance is still going to exacerbate. Thus, the techniques neglecting performance heterogeneity will directly lead to discrimination in downstream services for various recommendation and allocation, resulting in a double-edged-sword feature on information techniques. In this work, we devise a fairness-aware mobile computing technique, FairSTG, which advances the modeling of spatiotemporal heterogeneity on immediate observations to suppress the heterogeneity across prediction performances, empowering the intelligence of location-based applications with fairness. Specifically, our FairSTG takes node-level series as samples while learns the representations in a graph-based holistic perspective, allowing the flexibility to collaboratively transfer advantages of well-learned local representations to challenging ones with adaptive mix-up, thus alleviating the negative edge of technique and facilitating the sustainable urban computing. Our experiments show the effectiveness on applications from environments to traffic management, and further case studies demonstrate FairSTG can potentially alleviate the risks on traffic resource allocation for underrepresented urban regions. To this end, we believe our solution can be referential and generable to other mobile computing services, thus satisfying a broad audience in human-centered mobile computing techniques.

\section{Conclusion}
In this work, we uncover a new heterogeneity phenomenon designated prediction unfairness in spatiotemporal forecasting, and attribute such performance heterogeneity to data sufficiency and inherent regularity within observations. By organizing observations into node-level sequential samples, we then propose a FairSTG tailored for systematically mitigating sample-level unfairness in spatiotemporal graph learning tasks, where the core idea is to exploit well-learned advantageous node-level samples to help those samples with similar patterns but are difficult to learn. Our FairSTG consists of a spatiotemporal feature extractor for model warm-up and representation initialization, a fairness-aware learning architecture for actively identifying challenging samples and collaborative representation enhancement, and integrated fairness learning objective with fairness signal self-supervision and fairness constraints suppressing sample-level prediction heterogeneity. Substantial experimental results demonstrate that our FairSTG arrives the comparable performances with SOTA baselines, simultaneously significantly improving the quality of fairness guarantee. Our experiments verify FairSTG can effectively mitigate the spatial node-level heterogeneity by cross-sample enhancement from spatial perspective while suppress the temporal heterogeneity by retrieving compensatory samples from different graph steps. We believe our FairSTG, pays more attention to the urban fairness, can be a paradigm of learning-based urban computing, which suppresses the negative edge of information technology, from the promotion of fairness-aware techniques. For future work, we will further explore the root causes of prediction unfairness from both model and data aspects, and develop data-adaptive fairness learning to accommodate different datasets.

\section*{Acknowledgments}
This paper is partially supported by the National Natural Science Foundation of China (No.62072427, No.12227901), the Project of Stable Support  for Youth Team in Basic Research Field, CAS (No.YSBR-005), and the grant from State Key Laboratory of Resources and Environmental Information System.

% {\appendix[Proof of the Zonklar Equations]
% Use $\backslash${\tt{appendix}} if you have a single appendix:
% Do not use $\backslash${\tt{section}} anymore after $\backslash${\tt{appendix}}, only $\backslash${\tt{section*}}.
% If you have multiple appendixes use $\backslash${\tt{appendices}} then use $\backslash${\tt{section}} to start each appendix.
% You must declare a $\backslash${\tt{section}} before using any $\backslash${\tt{subsection}} or using $\backslash${\tt{label}} ($\backslash${\tt{appendices}} by itself
%  starts a section numbered zero.)}

% %{\appendices
% %\section*{Proof of the First Zonklar Equation}
% %Appendix one text goes here.
% % You can choose not to have a title for an appendix if you want by leaving the argument blank
% %\section*{Proof of the Second Zonklar Equation}
% %Appendix two text goes here.}

% \section{References Section}
% You can use a bibliography generated by BibTeX as a .bbl file.
%  BibTeX documentation can be easily obtained at:
%  http://mirror.ctan.org/biblio/bibtex/contrib/doc/
%  The IEEEtran BibTeX style support page is:
%  http://www.michaelshell.org/tex/ieeetran/bibtex/
 
%  % argument is your BibTeX string definitions and bibliography database(s)
% %\bibliography{IEEEabrv,../bib/paper}
% You can manually copy in the resultant .bbl file and set second argument of $\backslash${\tt{begin}} to the number of references
%  (used to reserve space for the reference number labels box).

% \bibliographystyle{IEEEtran}
% \bibliography{reference}
\newpage
\bibliographystyle{IEEEtran}
\bibliography{reference}

% \end{thebibliography}

\newpage

\section{Biography Section}
 

\begin{IEEEbiography}[{\includegraphics[width=1in,height=1.25in]{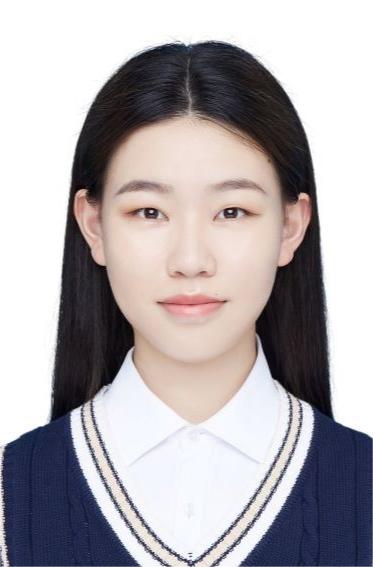}}]{Gengyu Lin} is currently pursuing a M.S. degree at University of Science and Technology of China. She obtained her B.S. degree from Beijing Normal University in 2022. Her mainly research interests include spatiotemporal data mining and fairness algorithms.
\end{IEEEbiography}
\vspace{-0.5in}

\begin{IEEEbiography}[{\includegraphics[width=1in,height=1.25in]{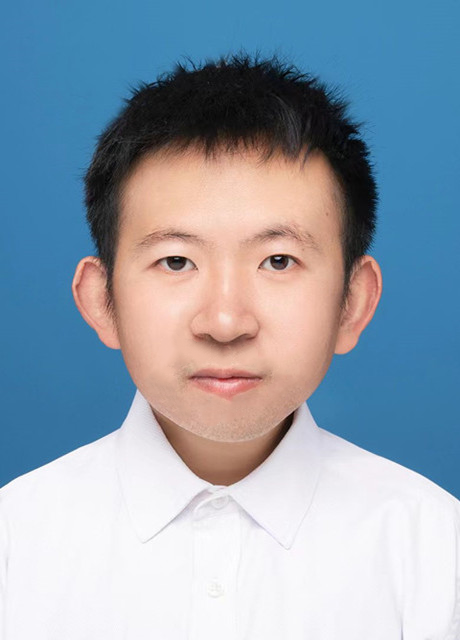}}]{Zhengyang Zhou} is now an associate researcher at Suzhou Institute for Advanced Research, University of Science and Technology of China (USTC). He got his Ph.D. degree at University of Science and Technology of China in 2023. He has published over 20 papers on top conferences and journals such as KDD, ICLR, TKDE, WWW, AAAI, SDM and ICDE. His mainly research interests include human-centered urban computing, and mobile data mining. 
\end{IEEEbiography}
\vspace{-0.5in}

\begin{IEEEbiography}[{\includegraphics[width=1in,height=1.25in]{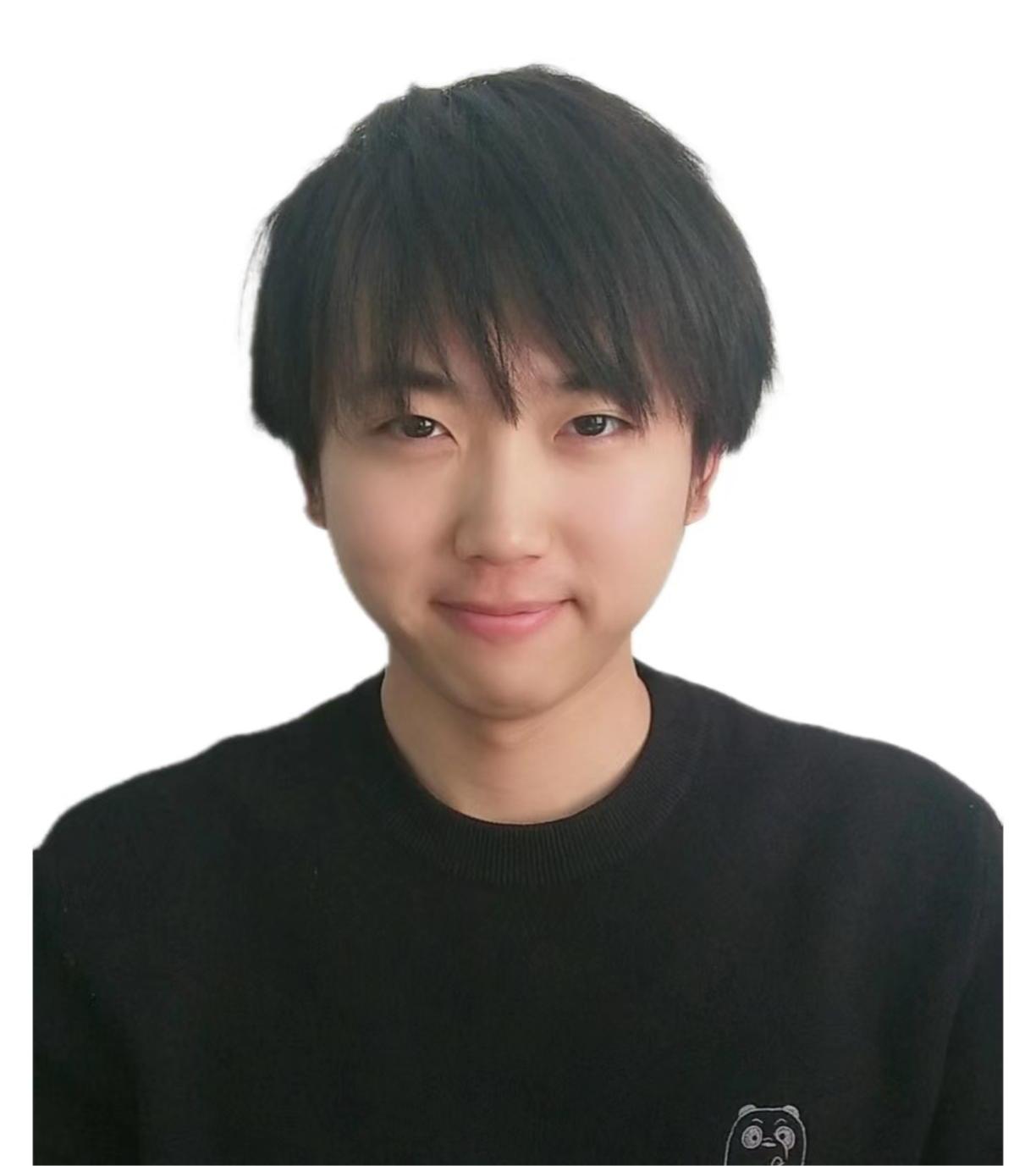}}]{Qihe Huang} is currently pursuing a M.S. degree at University of Science and Technology of China. He obtained his B.S. degree from Nanjing University of Information Science and Technology in 2022. His research centers on spatiotemporal data mining and mobile edge computing.
\end{IEEEbiography}
\vspace{-0.5in}

\begin{IEEEbiography}[{\includegraphics[width=1in,height=1.25in]{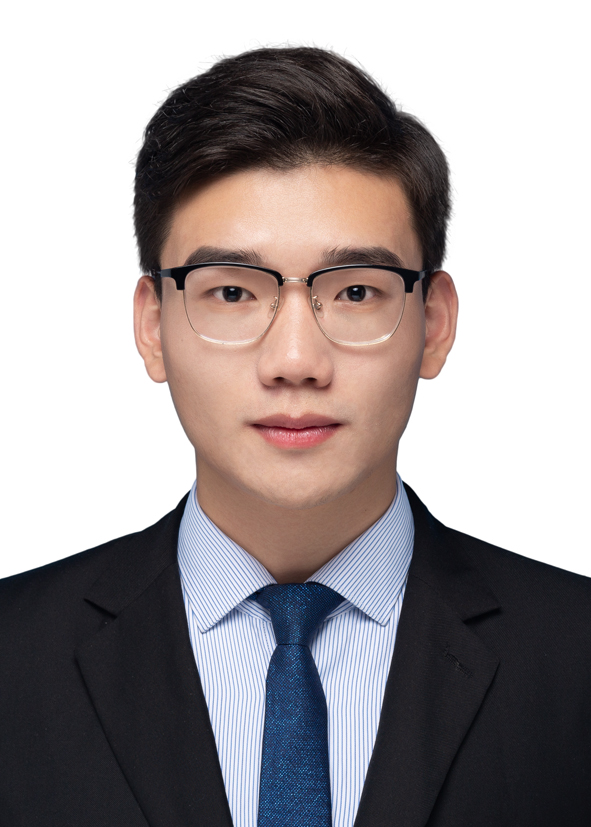}}]{Kuo Yang} received his BE degree from Northeastern University at Qinhuangdao, China, in 2021. He is now a  PhD  candidate at School of Data Science, USTC. His mainly research interests are data-driven urban management, spatiotemporal data mining and mobile computing. 
	\end{IEEEbiography}
	\vspace{-0.5in}

\begin{IEEEbiography}[{\includegraphics[width=1in,height=1.25in]{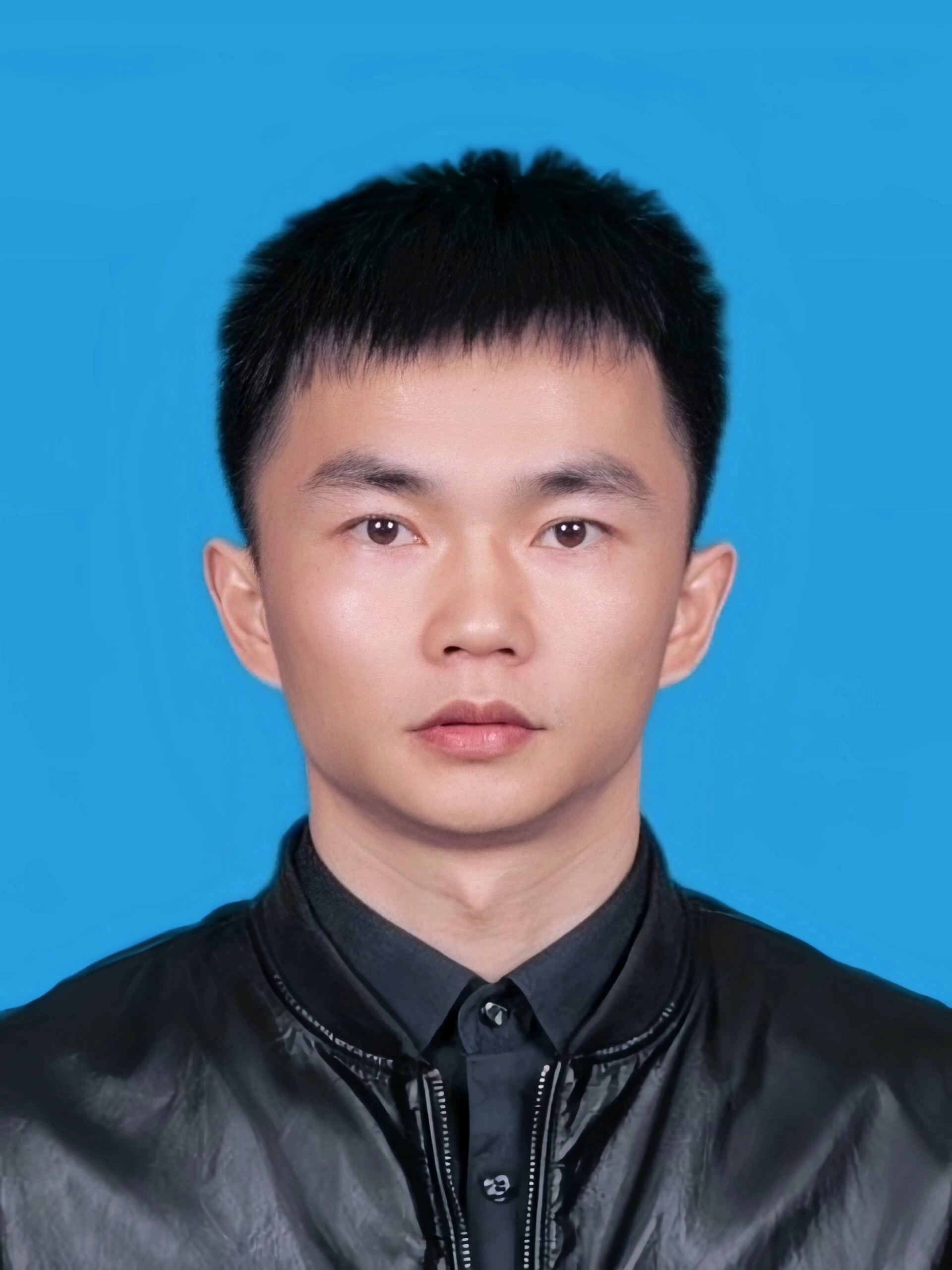}}]{Shifen Cheng} is an Associate Professor of State Key Laboratory of Resources and Environmental Information Systems, Institute of Geographical Sciences and Natural Resources Research, Chinese Academy of Sciences. He received his Ph.D. degree from Institute of Geographical Sciences and Natural Resources Research, Chinese Academy of Sciences. His research interests include spatiotemporal data mining, urban computing and intelligent transportation.
	\end{IEEEbiography}
	\vspace{-0.5in}

\begin{IEEEbiography}[{\includegraphics[width=1in,height=1.25in]{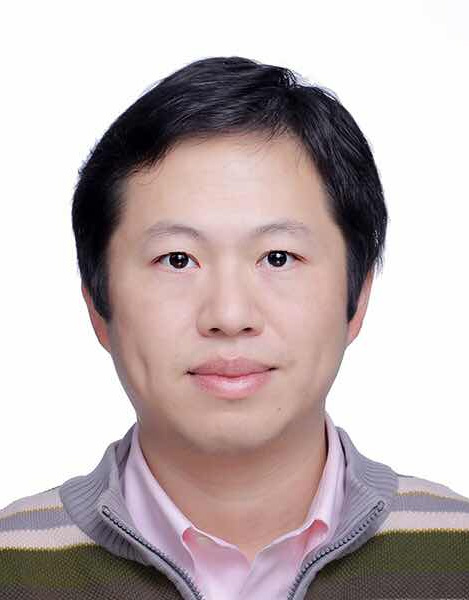}}]{Yang Wang} is now an associate professor at School of Computer Science and Technology, School of Software Engineering, and School of Data Science in USTC. He got his Ph.D. degree at University of Science and Technology of China in 2007. Since then, he keeps working at USTC till now as a postdoc and an associate professor successively. Meanwhile, he also serves as the vice dean of school of software engineering of USTC. His research interest mainly includes wireless (sensor) networks, distribute systems, data mining, and machine learning, and he is also interested in all kinds of applications of AI and data mining technologies especially in urban computing and AI4Science.
\end{IEEEbiography}
\vspace{-0.5in}

% \begin{IEEEbiography}[{\includegraphics[width=1in,height=1.25in,clip,keepaspectratio]{./figure/fig1.png}}]{Michael Shell}

% \end{IEEEbiography}



\vfill

\end{sloppypar}
\end{document}